\documentclass{article}

\PassOptionsToPackage{numbers, compress}{natbib}

    \usepackage[final]{neurips_2025}

\usepackage[utf8]{inputenc} 
\usepackage[T1]{fontenc}    
\usepackage{hyperref}       
\usepackage{url}            
\usepackage{booktabs}       
\usepackage{amsfonts}       
\usepackage{nicefrac}       
\usepackage{microtype}      
\usepackage{xcolor}         
\usepackage{graphicx}%
\usepackage{multirow}%
\usepackage{multicol}
\usepackage{amsmath,amssymb,amsfonts}%
\usepackage{amsthm}%
\usepackage{mathrsfs}%
\usepackage[title]{appendix}%
\usepackage{xcolor}%
\usepackage{textcomp}%
\usepackage{manyfoot}%
\usepackage{booktabs}%
\usepackage{algorithm}%
\usepackage{algorithmicx}%
\usepackage{algpseudocode}%
\usepackage{listings}%
\usepackage{caption} 

\usepackage{multirow}
\usepackage{array}
\usepackage{booktabs}
\usepackage{makecell}
\usepackage{enumitem}
\usepackage{titlesec}
\usepackage{xcolor}
\usepackage{pifont}

\newcommand{\cmark}{\textcolor{green!60!black}{\ding{51}}}
\newcommand{\xmark}{\textcolor{red}{\ding{55}}}

\author{
    Yuetong Fang\textsuperscript{\rm 1},~ 
    Deming Zhou\textsuperscript{\rm 1},~
    Ziqing Wang\textsuperscript{\rm 4},~
    Hongwei Ren\textsuperscript{\rm 1},~
    \\
    \textbf{Zecui Zeng}\textsuperscript{\rm 3},~
    \textbf{Lusong Li}\textsuperscript{\rm 3},~
    \textbf{Shibo Zhou}\textsuperscript{\rm 2}\thanks{Corresponding author},~
    \textbf{Renjing Xu}\textsuperscript{\rm 1}\footnotemark[1]~\\
    \\
    \textsuperscript{\rm 1}\small{The Hong Kong University of Science and Technology (Guangzhou)}~\\
    \textsuperscript{\rm 2}\small{Brain Mind Innovation INC}~
    \textsuperscript{\rm 3}\small{JD Explore Academy}~
    \textsuperscript{\rm 4}\small{Northwestern University}\\
    \\
    \small{\texttt{yfang870@connect.hkust-gz.edu.cn}\texttt{,}}\\   \small{\texttt{bob@brain-mind.com.cn}\texttt{,}~\texttt{renjingxu@hkust-gz.edu.cn}}
}

\begin{document}

\title{Spiking Neural Networks Need High-Frequency Information}

\maketitle

\begin{abstract}
  {Spiking Neural Networks promise brain-inspired and energy-efficient computation by transmitting information through binary (0/1) spikes. Yet, their performance still lags behind that of artificial neural networks, often assumed to result from information loss caused by sparse and binary activations. In this work, we challenge this long-standing assumption and reveal a previously overlooked frequency bias:  \textbf{spiking neurons inherently suppress high-frequency components and preferentially propagate low-frequency information}. This frequency-domain imbalance, we argue, is the root cause of degraded feature representation in SNNs. Empirically, on Spiking Transformers, adopting Avg-Pooling (low-pass) for token mixing lowers performance to 76.73\% on Cifar-100, whereas replacing it with Max-Pool (high-pass) pushes the top-1 accuracy to 79.12\%. Accordingly, we introduce {\textbf{{Max-Former}}} that restores high-frequency signals through two frequency-enhancing operators: (1) extra Max-Pool in patch embedding, and (2) Depth-Wise Convolution in place of self-attention. Notably, \textbf{Max-Former} attains 82.39\% top-1 accuracy on ImageNet using only 63.99M parameters, surpassing Spikformer (74.81\%, 66.34M) by +7.58\%. Extending our insight beyond transformers, our \textbf{Max-ResNet-18} achieves state-of-the-art performance on convolution-based benchmarks: 97.17\% on CIFAR-10 and 83.06\% on CIFAR-100. We hope this simple yet effective solution inspires future research to explore the distinctive nature of spiking neural networks. \href{https://github.com/bic-L/MaxFormer}{Code is available: https://github.com/bic-L/MaxFormer}.
  }
\end{abstract}

\includegraphics[width=1\linewidth]{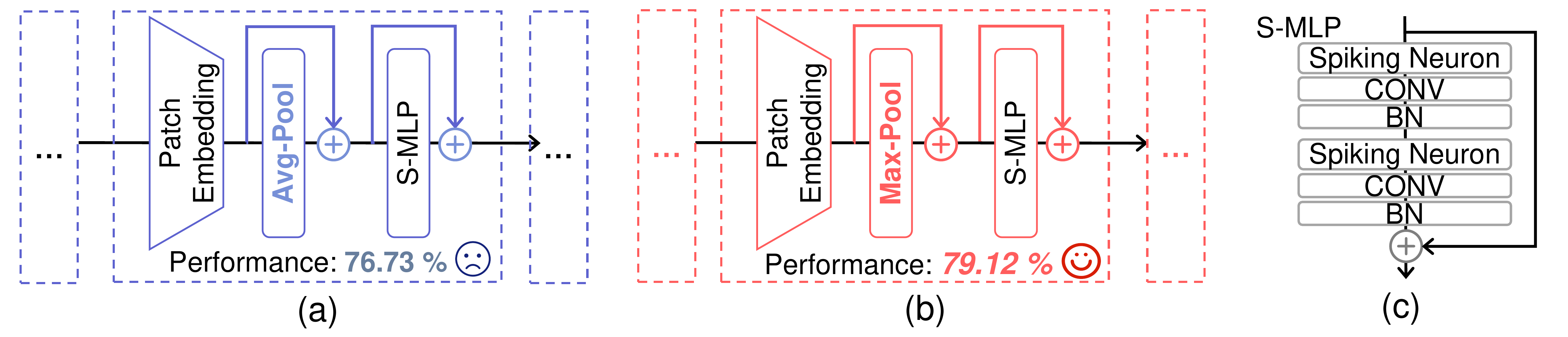}
\captionof{figure}{Spiking Transformer architectures: (a) Avg-Pool vs. (b) Max-Pool for token mixing, with (c) detailed implementation of the Spiking MLP (S-MLP) block. In mainstream (non-spiking) Vision Transformer research, Avg-Pool that captures global low-frequency patterns is a more common token mixing strategy than Max-Pool (high-pass)~\cite{yu2022metaformer, yu2023metaformer}. Surprisingly, in Spiking Transformers, replacing Avg-Pool with Max-Pool yields a +2.39\% improvement on Cifar-100.   }
\label{fig:fig1_cmp}

\section{Introduction}

Spiking neural networks (SNNs) are emerging as an energy-efficient alternative to conventional artificial neural networks (ANNs)~\cite{lecun2015deep, he2024diffusion}. Their efficiency arises from spiking neurons that utilize spatiotemporal dynamics to mimic biological computation in the human brain~\cite{roy2019towards}. In ANNs, all neurons within the same layer must await the complete processing of real-valued, dense tensors before any information can flow to the subsequent layer. SNNs, however, transmit information asynchronously, with spiking neurons consuming energy only when receiving or emitting spikes (``1''), otherwise remaining inactive~\cite{akopyanTruenorthDesignTool2015, davies2018loihi}. This binary activation pattern enables SNNs to replace the energy-intensive multiply-and-accumulate (MAC) operations that are essential in ANNs with much simpler spike-based accumulation. Leveraging the energy-efficiency benefits, modern SNN variants, such as Spiking Transformers that integrate the powerful Transformer architecture with spike-based computing, have gained growing attention~\cite{zhou2022spikformer, wang2023masked}.\vspace{5.5pt}

\begin{figure*}[t!]
\includegraphics[width=1.0\linewidth]
{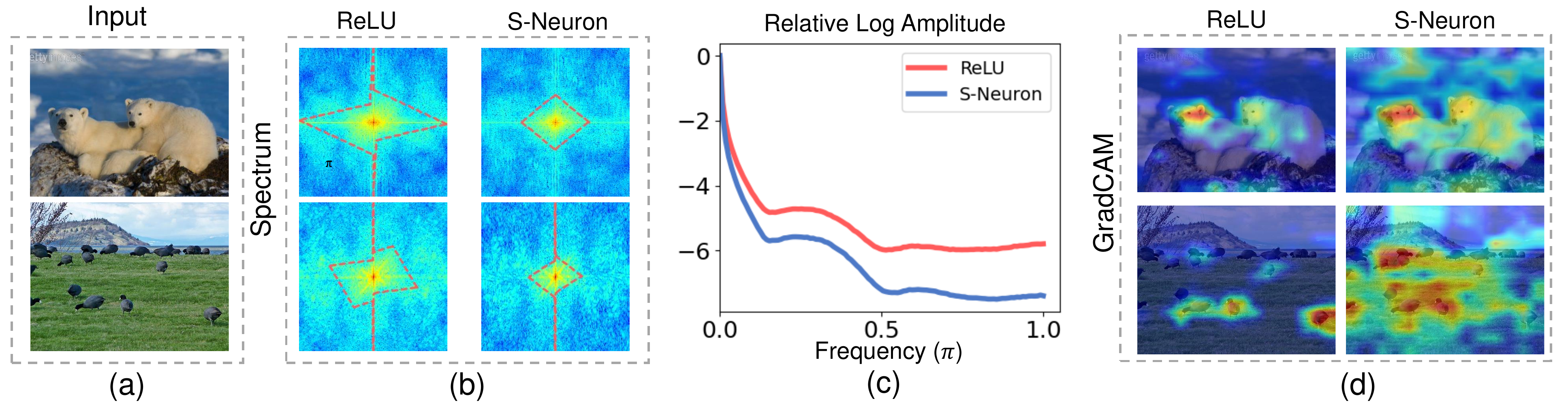}
\caption{ Comparison between ReLU and spiking neuron (S-Neuron): (a) Input images; (b) Fourier spectrum analysis of output features processed as input$\rightarrow$ activation$\rightarrow$ weighting, with high-frequency regions marked (red dashed boxes: regions >0.55$\times$ max amplitude) and (c) the corresponding relative log amplitude; (d) GradCAM comparison with identical architectural setting following~\cite{wang2023masked}, with the converted Spiking Transformer using 256 timesteps. Spiking neurons cause the rapid dissipation of high-frequency components, which consequently leads to the degradation of feature representations. }
\vspace{-0.7cm}
\label{fig:freq_cmp}
\end{figure*}

Despite their energy efficiency, the discrete nature of spike-based computation presents both opportunities and challenges. A major obstacle for SNNs remains their performance gap relative to ANNs. This gap is often attributed to the so‑called ``representation error''~\cite{liDifferentiableSpikeRethinking2021, guo2022loss, dingOptimalAnnsnnConversion2021}, which argues binary spike trains inherently limit the precision of feature representations compared to continuous activations. However, this seems inconsistent with the established consensus in the standard deep learning literature, where low-bit and even binary networks can still achieve comparable accuracy~\cite{li2022q, rastegari2016xnor}. Further, it should be noted that SNNs operate on temporal sequences: while spiking neurons strictly transmit binary signals at each individual time step, a train of spikes spanning $n$ simulation timesteps can encode activation values with at least $\log(n)$-bit precision~\cite{adrian1926impulses, liFreeLunchANN2021}. The conflicting observation reveals an unexplored dimension in understanding SNN performance limitations.
\vspace{5.5pt}

It is natural to think from the frequency domain. Spiking neurons produce discrete, pulse-like activations, fundamentally distinguishing their frequency response from continuous activation functions commonly used in standard networks (e.g., ReLU~\cite{lecun1998gradient}). Prior works have suggested that spiking neurons may enrich signals with local details (high frequencies)~\cite{rombouts2010fractionally, fang2024spiking}. However, in Figure~\ref{fig:freq_cmp} (b-c), we observed a surprising phenomenon: when examining the end-to-end information flow of input $\rightarrow$ activation $\rightarrow$ weighting, rather than focusing solely on the property of activation functions, spiking neurons tend to propagate low-frequency information more prominently than ReLU.
\vspace{5.5pt}

Feature degradation observed in SNNs may instead originate from the rapid dissipation of high-frequency components, which prevents the network from effectively capturing local, fine-grained information (Figure~\ref{fig:freq_cmp} (d)). To support this finding, we perform a simple experiment in which non‑parametric pooling operators, i.e., Max-Pool and Avg-Pool, serve as token mixers in Spiking Transformers, shown in Figure~\ref{fig:fig1_cmp}. From the frequency domain perspective, Max-Pool excels at capturing local high-frequency details (e.g., local edges/textures), whereas Avg-Pool favors global low-frequency patterns. Intriguing, Spiking Transformers exhibit opposite preferences to ANNs in token mixing:  while standard Transformers typically employ Avg-Pool for token mixing~\cite{yu2022metaformer,yu2023metaformer}, replacing it with Max-Pool in Spiking Transformers yields a +2.39\% improvement on CIFAR-100, making it surpass the well-tuned Spikformer~\cite{zhou2022spikformer} baseline by 0.97\%.  
\vspace{5.5pt}

Overall, this work provides further theoretical and empirical evidence supporting the view that high‑frequency information is essential for SNNs:
\begin{itemize}[leftmargin=1em, itemsep=0em, labelsep=0.5em]
\item  We provide the first theoretical proof that spiking neurons inherently act as low‑pass filters at the network level, revealing their tendency to suppress high‑frequency features.

\item  We propose {\textbf{{Max-Former}}}, which restores high‑frequency information in Spiking Transformers via two lightweight modules: extra Max-Pool in patch embedding and Depth-Wise Convolution (DWC) in place of early-stage self-attention.

\item Restoring high-frequency information significantly improves performance while saving energy cost. On ImageNet, {\textbf{Max-Former}} achieves 82.39\% top-1 accuracy (+7.58\% over Spikformer) with 30\% energy consumption and lower parameter count (63.99M vs. 66.34M).

\item Extending the insight beyond transformers, \textbf{Max-ResNet-18} achieves state-of-the-art performance on convolution-based benchmarks: 97.17\% on CIFAR-10 and 83.06\% on CIFAR-100.
\end{itemize}
 
We believe this straightforward yet powerful solution will motivate future research to explore the unique properties of SNNs, beyond the established practice in standard deep learning.


\section{Preliminary and Related Works}

\subsection{Spiking Neuron Models}

SNNs implement spike-driven processing through biologically-inspired neuron models for non-linear activations. The Leaky Integrate-and-Fire (LIF) model is a widely adopted abstraction of this behavior, offering an effective balance between biological plausibility and computational efficiency~\cite{maass1997networks}. The discretized LIF model under each simulation timestep $n$ can be formulated as:

\hspace{-0.7cm}
\begin{minipage}{0.45\textwidth}
\begin{align}
U[n] &= f(V[n-1], I[n]), \label{eq:u} \\
S[n] &= H\bigl(U[n] - V_{\mathrm{th}}\bigr), \label{eq:s}
\end{align}
\end{minipage}
\hfill
\begin{minipage}{0.55\textwidth}
\begin{align}
V[n] &=
\begin{cases}
U[n] -V_{\mathrm{th}}, & S[n]=1,\\
U[n],      & S[n]=0,
\end{cases} \label{eq:v}
\end{align}
\end{minipage}
\vspace{5.5pt}

where $\beta$ is the decay factor, $V_{\mathrm{th}}$ is the firing threshold, and $H(\cdot)$ refers to the Heaviside step function that determining spike generation: $S[n] = H\bigl(U[n] - V_{\mathrm{th}}\bigr))$ = 1 when $U[n] \geq V_{\mathrm{th}}$, otherwise remains inactive ($S[n]$ = 0). The charging process of the LIF neuron is determined by $f(\cdot)$:
\begin{align}
f(V[n-1], I[n]) &= \beta V[n-1] + (1 - \beta) I[n] \label{eq:lif}
\end{align}
At each timestep $n$, the current membrane potential $U[n]$ is updated by integrating the time-domain signal $I[n]$ corresponding to input data or intermediate operations like Conv and MLP. If $U[n]$ exceeds the threshold $V_{\text{th}}$, the neuron fires a spike ($S[n]$ = 1). $V[n]$ records membrane potential over time given the decay factor $\beta$ and output spike activity. If the neuron does not fire, then $V[n]$ = $U[n]$. Notably, the LIF model simplifies to integrate-and-fire (IF) neurons when eliminating the membrane potential decay process between timesteps. Its charging process can be formulated as:
\begin{align}
f(V[n-1], I[n]) = V[n-1] + I[n] \label{eq:if}
\end{align}

\subsection{Spiking Neural Networks}
Drawing inspiration from biological neurons, SNNs extend conventional ANNs by incorporating temporal dynamics and discrete spike-based communication~\cite{roy2019towards}. Leveraging this spike-driven mechanism, neuromorphic chips implement computation through event-driven spike routing and accumulation, which substitutes energy-intensive matrix–vector multiplications~\cite{davies2018loihi, akopyanTruenorthDesignTool2015}. This facilitates high parallelism, scalability, and exceptional power efficiency, with power consumption typically in the range of tens to hundreds of milliwatts~\cite{basu2022spiking}.
\vspace{5.5pt}


 Recently, the development of modern SNNs, e.g., Spiking Transformer, has demonstrated both attractive performance and reduced energy consumption~\cite{zhou2022spikformer, wang2023masked, fang2025dynamic}. Spikformer~\cite{zhou2022spikformer} pioneered the spike-based self-attention mechanism called Spiking Self Attention (SSA) that utilizes sparse spike-form Query, Key, and Value vectors to eliminate the need for energy-intensive softmax operations.  Following its success, many works endeavor to enhance Spiking Transformers by adapting advanced ANN Transformer architectures~\cite{yao2024spike, zhou2024qkformer} or devising complicated spike coding mechanisms to reduce representation error (e.g., multi-threshold~\cite{hao2024lm}/~multi-spike neurons~\cite{wang2025adaptive, yao2025scaling}). \vspace{5.5pt}
 
 Here, we instead address a fundamental question: What truly limits SNNs' performance compared to ANNs? Our investigation reveals that the answer lies in frequency properties -— specifically, that spiking neurons function as low-pass filters, impeding the propagation of high-frequency detail within the network. In addition to theoretical proof, we probe the importance of high-frequency information through our {{Max-Former}}, which features two frequency-enhancing operators: Max-Pool in patch embedding and DWC in place of early-stage self-attention. We further validate this principle in convolutional architectures with our proposed Max‑ResNet.

\section{Methods}

In this section, we first present a theoretical analysis of the frequency properties of spiking neurons. We show that, although the raw output spike trains of spiking neurons appear spectrally all‑pass due to their impulse‑shaped spike waveform, the resulting high‑frequency components are merely superficial and cannot be propagated through the network. In fact, spiking neurons act as low‑pass filters at the network level. This is a fundamental problem that has been overlooked in previous works. Building on this insight, we probe the importance of high-frequency information in SNNs through Max-Former, which strategically employs high-pass operators (Max-Pool and DWC) to restore high-frequency details and avoid feature degradation.
\vspace{5.5pt}

\begin{figure*}[ht]
\vspace{-0.1cm}
\includegraphics[width=1\linewidth]
{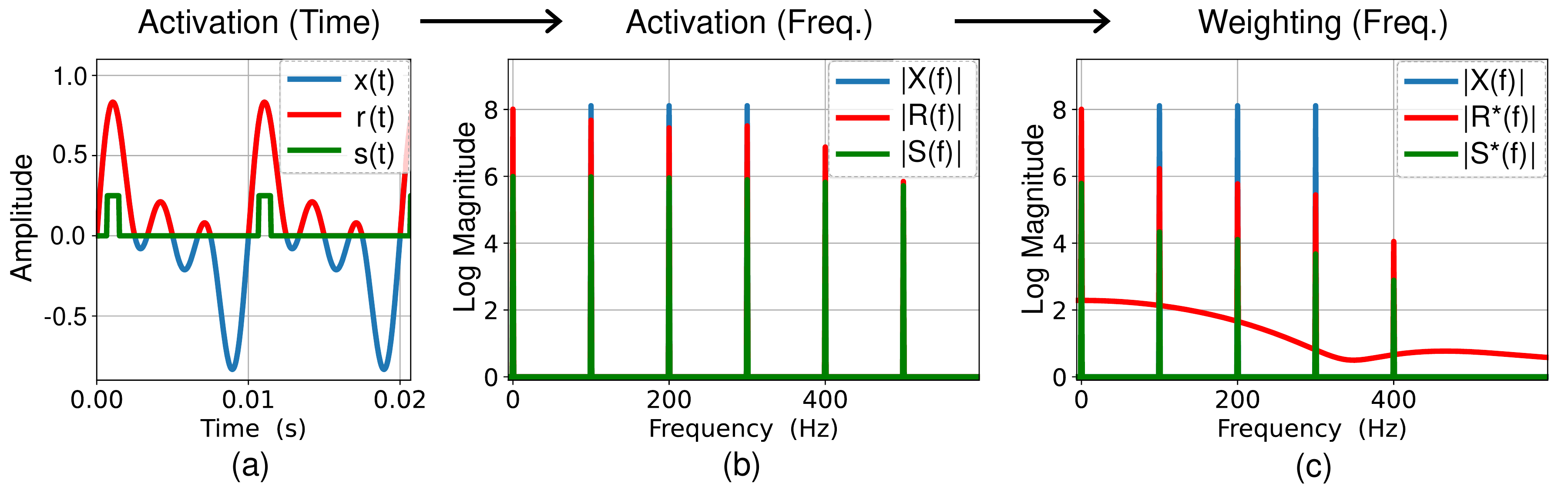}
\vspace{-0.6cm}
\caption{  Time-frequency analysis of ReLU and spiking neurons.
(a) Time-domain signals of input $x(t) = \frac{1}{3}(\sin(2\pi \cdot 100t) + \sin(2\pi \cdot 200t) + \sin(2\pi \cdot 300t))$ (blue), ReLU-processed: $r(t)$ (red), spiking output of a LIF neuron with the $\beta = 0.25$: $s(t)$ (green). (b) Fourier analysis of $x(t)$, $r(t)$, and $s(t)$. (c) Fourier analysis of linear transformed (CONV/MLP) activations, where ReLU expands the frequency bandwidth of the input signal, while the spiking neuron shows high-frequency attenuation.
 }
 \vspace{-0.4cm}

\label{fig:freq_analysis}
\end{figure*}

\subsection{Spiking Neurons are Low-pass Filters}

We begin with an intuitive time-frequency analysis using an input $x(t) = \frac{1}{3}(\sin(2\pi \cdot 100t) + \sin(2\pi \cdot 200t) + \sin(2\pi \cdot 300t))$ as shown in Figure \ref{fig:freq_analysis}. The results reveal three key observations: (1) In the time domain, the ReLU output ($r(t)$) perfectly follows $x(t) > 0$ while spiking neurons selectively respond to 100Hz (Figure~\ref{fig:freq_analysis} (a)); (2) However, the spiking outputs' spectral response |S(f)| still appears nearly all‑pass, which contradicts the low‑frequency behavior observed in the time domain. These spurious high‑frequency components actually arise from the impulse‑shaped spike waveform itself rather than from genuine (Figure \ref{fig:freq_analysis} (a-b)); (3) The waveform‑induced high‑frequency components cannot be propagated across layers, resulting in a network‑level low‑pass behavior. When considering the whole process from input $\rightarrow$ activation $\rightarrow$ linear transform, ReLU expands the frequency bandwidth of $x(t)$~\cite{kechris2024dc}, whereas the spiking neuron exhibits strong high-frequency attenuation (Figure~\ref{fig:freq_analysis} (c)).  
\vspace{5.5pt}

We first examine the charging process of spiking neurons to theoretically analyze their frequency-selective properties. Given Equ.~\eqref{eq:v} and Equ.~\eqref{eq:lif}, this can be formulated as:
\begin{equation}
V[n] = \beta V[n-1] + (1 - \beta)I[n]
\label{eq:lif_update_t}
\end{equation}
 Applying the Z–transform with \(\mathcal{Z}\{V[n-1]\}=z^{-1}V(z)\) yields:
\begin{equation}
V[z] = \beta z^{-1}V[z] + (1 - \beta)I[z]
\label{eq:lif_update_z}
\end{equation}
which can be rearranged to formulate the transfer function from input current to membrane potential:
\begin{equation}
H(z) = \frac{V(z)}{I(z)} = \frac{1 - \beta}{1 - \beta z^{-1}} ,\quad 0 \leq \beta < 1
\label{eq:transfer_function}
\end{equation}
Equ.~\eqref{eq:transfer_function} is exactly the form of a first‐order infinite‐impulse‐response (IIR) low‐pass filter with a single pole at \(z=\beta\).  Accordingly, as \(\beta\), i.e., the pole, approaches 1, LIF neurons exhibit stronger low‐frequency selectivity. Notably, the decay factor $\beta$ correlates to the membrane time constant $\tau$ as $\beta = 1 - \tfrac{1}{\tau}$, with $\tau$ ranges from $1$ to $+\infty$, a smaller $\tau$ yields a smaller $\beta$. From an intuitive standpoint, a shorter time constant allows the membrane potential to respond within narrower temporal windows, rendering the neuron more sensitive to higher‑frequency inputs. \vspace{5.5pt}

From the individual neuron to the network-level information transmission, the average membrane potential is positively and consistently correlated with the possibility of spike firing during operation. We approximate the inherently nonlinear spike‐generation process as linear around the firing threshold \(V_{th}\). Denoting the firing rate by \(\mathit{fr}(V)\), we define the local gain $k$ and approximate the Z–domain spike train \(S(z)\) by:

\vspace{-0.2cm}
\begin{minipage}{0.45\textwidth}
\begin{align}
k \;=\;\left.\frac{\partial\,\mathit{fr}}{\partial V}\right|_{V=V_{th}}
\label{eq:firing_sensitivity}
\end{align}
\end{minipage}
\hfill
\begin{minipage}{0.55\textwidth}
\begin{align}
S(z)\;\approx\;k\,V(z).
\label{eq:spike_train}
\end{align}
\end{minipage}


When output spike trains are weighted by a causal synaptic kernel \(w[n]\), the Z-transformed output current (\(y[n]=w[n]*s[n]\)) can be denoted as \(Y(z)=W(z)\,S(z)\). The overall input‐to‐output transfer function is obtained by combining this with \eqref{eq:transfer_function} and \eqref{eq:spike_train}:
\begin{equation}
H'(z)
\;=\;
\frac{Y(z)}{I(z)}
\;=\;S(z)\,W(z)\,H(z)\
\;=\;k\;W(z)\;\frac{1-\beta}{1-\beta\,z^{-1}}.
\label{eq:overall_transfer}
\end{equation}

The first-order low-pass IIR characteristics of $H(z)$ make the system $Y(z)$ inherently favor low-frequency signal components, regardless of whether the synaptic kernel $W(z)$ or the spike coding process $S(z)$ changes the gain or phase response. The low-pass term  $(\frac{1-\beta}{1-\beta z^{-1}})^L$  further amplifies the system's frequency selectivity when the process $H'(z)$ is cascaded $L$ times (layers). The complete formula is as follows:
\vspace{-0.2cm}
\begin{equation}
    H'_L(z)
    =\frac{Y_L(z)}{I(z)}
    =\prod_{i=1}^L \bigl[S_i(z)\,W_i(z)\,H(z)\bigr]
    =\Bigl(\prod_{i=1}^L k_i\,W_i(z)\Bigr)\,
     \Bigl(\frac{1-\beta}{1-\beta z^{-1}}\Bigr)^L
    \label{eq:overall_transfer_N}
\end{equation}
In the special case of the non‐leaky IF neurons, which obey the charging process in \eqref{eq:if}, $H(z)$ is formulated as:
\vspace{-0.2cm}
\begin{equation}
H_(z)
\;=\;
\frac{1}{1-z^{-1}},
\label{eq:if_transfer}
\end{equation}
 This corresponds to an ideal discrete-time low-pass filter with a pole at \(z=1\), which can yield a consistent conclusion with our previous analysis.


\subsection{Max-Former}
It remains unclear whether high‑frequency information is truly important for SNNs and whether restoring it can improve performance. Therefore, we systematically investigate the low-pass filtering characteristics of spiking neurons through Max-Former. To decouple frequency effects from model complexity, we: (1) replace self-attention with high-frequency-preserving DWC in the early stages, and (2) add Max-Pool in patch embedding to compensate spiking neurons' low-pass preference. Notably, compared to the quadratic computational complexity of self-attention, DWC and Max-Pool only require linear complexity relative to the sequence length and are more parameter-efficient. We consistently adopt the LIF neuron model throughout this work.

\begin{figure*}[ht]
\vspace{-0.1cm}
\includegraphics[width=1\linewidth]
{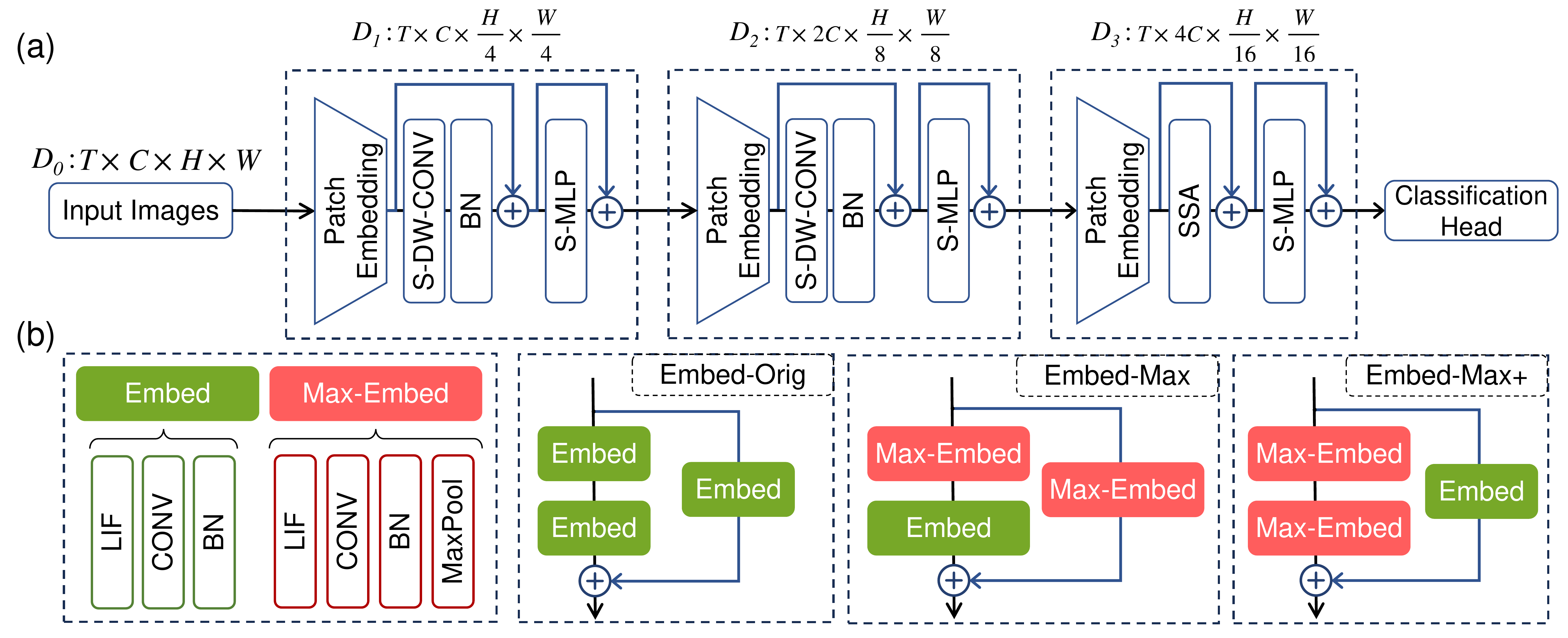}
\caption{  (a) Overview of Max-Former: we restore high-frequency signals by using lightweight DWCs instead of self-attention in the early stages. Following the hierarchical design of ~\cite{liuSwinTransformerHierarchical2021}, Max-Former adopts a 3-stage architecture. $D_i$: feature dimensions of stage-$i$.  (b) In  Max-Former's patch embedding stage, we propose three configurations (Embed-orig, Embed-Max, and Embed-Max+) to enhance high-frequency components. 
 }
 \vspace{-0.4cm}
 \label{fig:overview}
 \end{figure*}
 
\subsubsection{Overall Architecture}
Figure~\ref{fig:overview}(a) illustrates the overall framework of Max-Former. The architecture consists of 3 stages with $\frac{H}{4}\times\frac{W}{4}$, $\frac{H}{8}\times\frac{W}{8}$, and $\frac{H}{16}\times\frac{W}{16}$ tokens respectively, where $H$ and $W$ denote the height and width of the input image. Critically, MaxFormer processes information through discrete spikes over time. This spike-driven computing paradigm supports two types of input: \vspace{5.5pt}

\noindent(1) Event Streams:  Asynchronous events $e = [x, y, t, p]$ containing spatial coordinates $(x,y)$, timestamp $t$, and polarity $p$ are converted to event frames through temporal binning. Given original resolution $dt_o$ and target $dt = \alpha dt_o$, events are aggregated over $\alpha$ consecutive bins:
\vspace{-0.2cm}
\begin{equation}
I_t = \sum_{k=\alpha t}^{\alpha(t+1)-1} S_k \in \mathbb{R}^{2\times h \times w}
\end{equation}
\vspace{-0.3cm}

\noindent where $S_k$ denotes raw event data. The whole process denoises raw events and converts them to frame sequences at target temporal resolutions. \vspace{3.5pt}

\noindent(2) Static Images:  Conventional images are converted to spike sequences by: 1) Repeating static frames $T$ times, 2) Encoding pixel intensities to spikes using spiking neurons. The resulting input is formulated as: $I = \text{Spiking\_Embed}(\{I_t\}_{t=1}^T)$, which contains identical information across timesteps.

\subsubsection{Patch Embedding}

To transform the input into a tokenized representation, given input $\{S\} \in \mathbb{R}^{T \times C \times H \times W}$, the process of patch embedding is formulated as:
\vspace{-0.2cm}
\begin{equation}
\mathbf{Y} = \left(\mathcal{G}_1({\{S\}}) + {G}_2({\{S\}}\right),\quad \mathbf{Y} \in  \mathbb{R}^{T \times C' \times H' \times W'}
\label{eq:transformation}
\end{equation}
where $C' = 2C$, and $H' = \lfloor H/P \rfloor$, $W' = \lfloor W/P \rfloor$ with the patch size $P=4$.  To address spiking neurons' inherent frequency preference, we present three patch embedding configurations as shown in Figure~\ref{fig:overview}(b) :
\vspace{-0.2cm}
\begin{align}
&\text{Embed-Orig~~~:}  (\mathcal{G}_1, \mathcal{G}_2) = \text{(Embed, Embed)} \\
&\text{Embed-Max~~~:}  (\mathcal{G}_1, \mathcal{G}_2) = \text{(Max-Embed, Embed)} \\
&\text{Embed-Max+ :}  (\mathcal{G}_1, \mathcal{G}_2) = \text{(Max-Embed, Max-Embed)}
\end{align}
\vspace{-0.2cm}
where:
$\text{Embed} \equiv \{\text{LIF - CONV - BN}\}$, and $\text{Max-Embed} \equiv \{\text{LIF - CONV - BN - MaxPool}\}$. \vspace{5.5pt}

\subsubsection{Token Mixing}

In Transformers, lower layers typically require more high-frequency details, while higher layers benefit from more global information~\cite{raghu2021vision, si2022inception}. Like biological vision, high-frequency details enable early stages to learn low-level features while progressively building local-to-global representations. Accordingly, we replace early-stage self-attention with DWC to preserve high frequencies essential for local feature learning. Given input embedding $\mathbf{Y} \in \mathbb{R}^{T \times C \times H \times W}$, the spiking DWC is defined as:
\vspace{-0.4cm}
\begin{align}
\mathbf{Z}_c(\mathbf{Y})[i] &= \text{LIF}(\sum_{j \in \Omega(i)} w_{c,j} \cdot \mathbf{Y}_c[j])
\end{align}
\vspace{-0.35cm}

where $\Omega(i)$ denotes the local neighborhood of position $i$, $w_{c,j}$ represents learnable convolution weights for channel $c$, and $X_c, Z_c \in \mathbb{R}^{T \times H \times W}$ is the input and output slice for channel $c$.
For the final stage, we implement token mixing via Spiking Self-Attention (SSA)~\cite{zhou2022spikformer}. The SSA computation follows:
\vspace{-0.4cm}
\begin{align}
    &\mathbf{Z} = \text{LIF}(\text{BN}(\mathbf{Y}\mathbf{W})), \quad \mathbf{Z} \in \{Q, K, V\} \label{eq:ssa_input} \\
    &\text{SSA}(\mathbf{Q}, \mathbf{K}, \mathbf{V}) = \text{LIF}(\mathbf{QK^T}\mathbf{V} \cdot s) \label{eq:ssa_output}
\end{align}
\vspace{-0.45cm}

where $\mathbf{Q}, \mathbf{K}, \mathbf{V} \in \mathbb{R}^{T \times N \times H \times W}$ are spike-form tensors generated by learnable linear layers, $s$ is a scaling factor. SSA eliminates floating-point multiplications, ensuring spike-driven compatibility.

\begin{figure*}[t!]
\hspace{0.7cm}
\includegraphics[width=0.9\linewidth]
{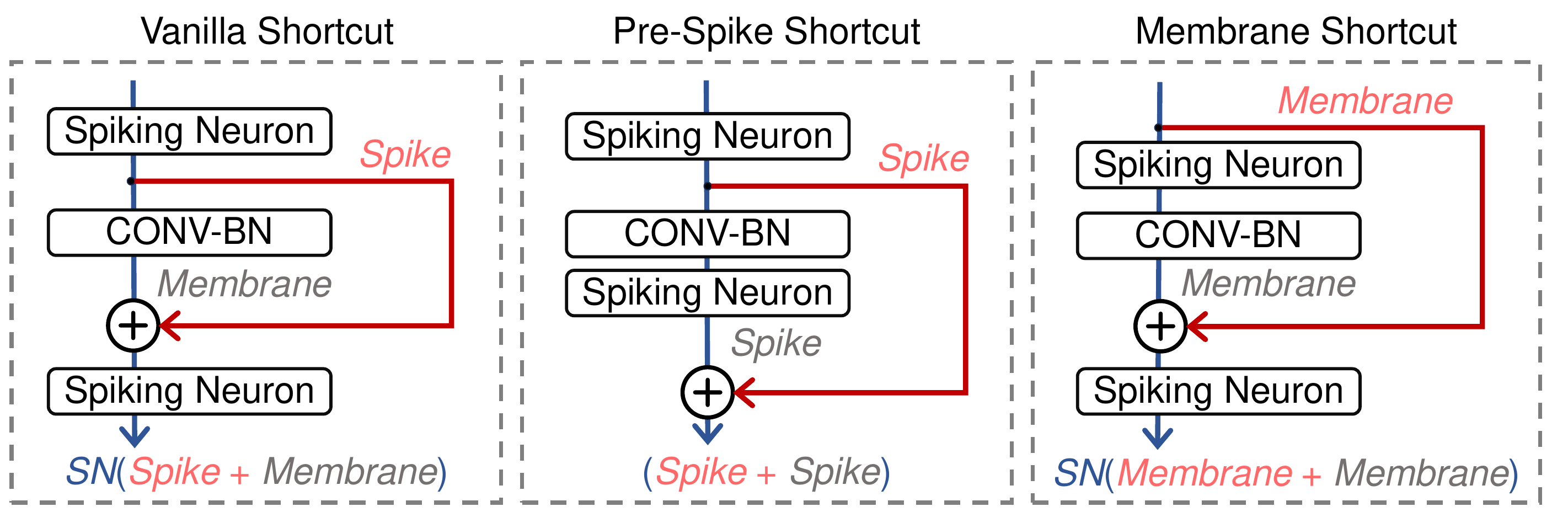}
\vspace{-0.2cm}
\caption{  Shortcut connection in SNNs. (Left) Vanilla Shortcut that combines spike and membrane potential. (Middle) Pre-Spike Shortcut that adds spike signals before neuron charging. (Right) Membrane Shortcut that directly connects membrane potentials, ensuring identical potential mapping while strictly preserving the spike-driven computing paradigm throughout the network.
 }
\vspace{-0.4cm}
 \label{fig:shortcut}
 \end{figure*}

\subsection{Membrane Shortcut}

Residual learning and shortcuts enable very deep networks to be trained effectively by providing identity shortcuts that preserve information flow and mitigate vanishing gradients~\cite{he2016identity}. For SNNs, a crucial consideration is maintaining compatibility with the spike-driven computing paradigm throughout all operations. As Fig.~\ref{fig:shortcut} shows, existing SNN shortcut implementations fall into three categories: (1) Vanilla Shortcut~\cite{zheng2021going}, (2) Pre-Spike Shortcut~\cite{fangDeepResidualLearning2021}, and (3) Membrane Shortcut~\cite{hu2024advancing}. \vspace{5.5pt}

The Vanilla Shortcut scheme~\cite{zheng2021going} directly connects spikes (binary) to membrane potentials (continuous), leading to a distribution mismatch that inherently violates the identity mapping principle. The Pre-Spike Shortcut~\cite{fangDeepResidualLearning2021} adds spike signals before neuron charging, resulting in summation values that range from 0 to 2. This disrupts the binary spike representation and spike-driven flow in SNNs. \vspace{5.5pt}

In this work, we adopt the Membrane Shortcut~\cite{hu2024advancing} for its dual advantages: it preserves identity mapping by directly connecting membrane potentials while maintaining binary spike outputs that remain inherently compatible with spike-driven computation. Unlike Vanilla or Pre-Spike Shortcuts, this approach ensures both mathematical consistency with residual learning principles and seamless compatibility with spike-driven operations. We provide a detailed analysis of its impact on model performance and energy costs in \textbf{Appendix~\ref{sec:a4}}.\vspace{-0.2cm}

\section{Experiment}

\begin{figure*}[ht]
\hspace{0.7cm}
\vspace{-0.1cm}
\includegraphics[width=0.9\linewidth]
{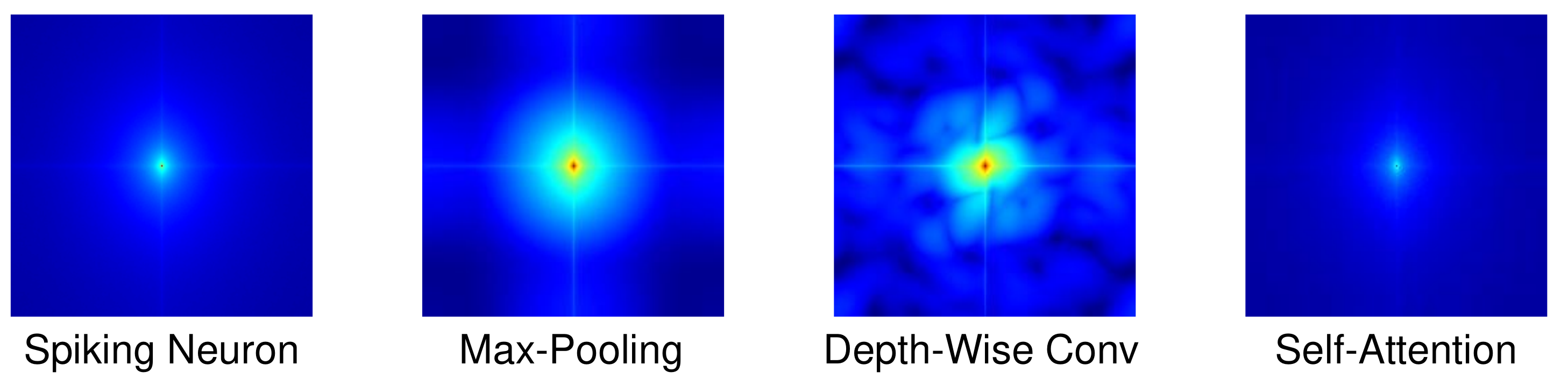}
\vspace{-0.0cm}
\caption{  Fourier spectrum of Spiking Neurons, Spiking Max-Pool, Spiking Depth-Wise Convolution and Spiking Self-attention. 
 }
 \vspace{-0.2cm}
 \label{fig:freq_show}
 \end{figure*}

 As shown in Figure~\ref{fig:freq_show}, Max-Former restores high-frequency information by grafting the merits of frequency-enhancing operators: Max-Pool and DWC. To empirically probe the importance of high-frequency information in Spiking Transformers, we evaluate Max-Former through comprehensive experiments on static datasets (CIFAR-10~\cite{lecunGradientbasedLearningApplied1998a}, CIFAR-100~\cite{krizhevskyLearningMultipleLayers2009a} and ImageNet~\cite{dengImagenetLargescaleHierarchical2009}) and neuromorphic datasets (CIFAR10-DVS~\cite{liCIFAR10DVSEventStreamDataset2017}, DVS128 Gesture~\cite{amir2017low}), with architecture configurations detailed in Table~\ref{tab:configs}. In addition, we design Max‑ResNet to further investigate the effect of high‑frequency restoration in convolutional architectures (model implementation is detailed in \textbf{Appendix~\ref{sec:max-resnet}}).  Experimental settings and energy analysis methods are detailed in \textbf{Appendix~\ref{sec:a1}} and\textbf{~\ref{sec:a2}}.

 \begin{table}[ht]
\vspace{-0.4cm}
  \centering
  \caption{Max-Former architecture configurations for different classification tasks. Notation DWC-$N$ denotes depth-wise convolution with kernel size $N \times N$.  For block settings: CIFAR-10/100: 3 stages (1/1/2 blocks); ImageNet: 3 stages (1/3/7 blocks); Neuromorphic: 2 stages (1/1 blocks).}
  \label{tab:configs}

  {
   \setlength{\tabcolsep}{3pt}    
    \setlength{\extrarowheight}{0pt}
    \renewcommand{\arraystretch}{0.85}

   \resizebox{\textwidth}{!}{%
     \begin{tabular}{ccccccc}
       \toprule
        \multirow{2}{*}{Dataset}
         & \multicolumn{2}{c}{Stage 1}
         & \multicolumn{2}{c}{Stage 2}
         & \multicolumn{2}{c}{Stage 3} \\
       \cmidrule(lr){2-3} \cmidrule(lr){4-5} \cmidrule(lr){6-7}
         & Patch Embed & Token Mix
         & Patch Embed & Token Mix
         & Patch Embed & Token Mix \\
       \midrule
       Cifar10~\cite{lecunGradientbasedLearningApplied1998a}/100~\cite{ krizhevskyLearningMultipleLayers2009a} & Embed-Orig & Identity & Embed-Max & DWC-3 & Embed-Max & SSA \\
       ImageNet~\cite{dengImagenetLargescaleHierarchical2009}         & Embed-Orig & DWD-7    & Embed-Max & DWC-5 & Embed-Max & SSA \\
       Neuromorphic~\cite{liCIFAR10DVSEventStreamDataset2017, amir2017low}     & Embed-Max+ & DWC-3    & Embed-Max & SSA   & \textemdash        & \textemdash  \\
       \bottomrule
     \end{tabular}%
   }
  }
  \vspace{0.0cm}

\end{table}
 
\subsection{Results on CIFAR and Neuromorphic Datasets}

\begin{table}[b!]
    \vspace{-0.5cm}
  \centering
  \caption{Performance Comparison on CIFAR10~\cite{lecunGradientbasedLearningApplied1998a}, CIFAR100~\cite{krizhevskyLearningMultipleLayers2009a}, DVS128~\cite{amir2017low} and CIFAR10-DVS~\cite{liCIFAR10DVSEventStreamDataset2017}. Param.: Parameter (M); Acc.:Top-1 Accuracy (\%); $T$: Simulation Timestep. \textsuperscript{*}Models trained from scratch using identical configurations. }
  \vspace{0.15cm}
  \label{tab:comparison}

  \resizebox{\textwidth}{!}{%
  {\scriptsize
   \setlength{\tabcolsep}{4pt}   
   \renewcommand{\arraystretch}{0.85} 

   \begin{tabular}{c*{12}{c}c}
     \toprule
    \multirow{2}{*}{Method}
     & \multicolumn{3}{c}{CIFAR10}
     & \multicolumn{3}{c}{CIFAR100}
     & \multicolumn{3}{c}{DVS128}
     & \multicolumn{3}{c}{CIFAR10-DVS} 
     & \multirow{2}{*}{\makecell{Membrane\\Shortcut}}\\
     \cmidrule(lr){2-4} \cmidrule(lr){5-7} \cmidrule(lr){8-10} \cmidrule(lr){11-13}
       & Param. & $T$ & Acc.
       & Param. & $T$ & Acc.
       & Param. & $T$ & Acc.
       & Param. & $T$ & Acc. \\
    \midrule
    ResNet-19 (ANN)~\cite{zhou2024qkformer}
       & 12.63 & 1 & 94.97
       & 12.63 & 1 & 75.35
       & \textemdash & \textemdash & \textemdash
       & \textemdash & \textemdash & \textemdash & \textemdash\\
     Max-Former (ANN)
       & 6.57  & 1 & 96.82
       & 6.60  & 1 & 82.41
       & \textemdash & \textemdash & \textemdash
       & \textemdash & \textemdash & \textemdash& \textemdash\\
     \midrule
     Spikformer~\cite{zhou2022spikformer}
       & 9.32 &  4 & 95.51
       & 9.32 &  4 & 78.21
       & 2.57 & 16 & 98.3
       & 2.57 & 16 & 80.9 & \xmark\\
     S-Transformer~\cite{yao2023spike}
       & 10.28 &  4 & 95.60
       & 10.28 &  4 & 78.40
       & 2.57  & 16 & {99.3}
       & 2.57  & 16 & 80.0 &\cmark\\
     SWformer~\cite{fang2024spiking}
       & 7.51 &  4 & 96.10
       & 7.51 &  4 & 79.30 
       & -  & - & -
       & 2.05  & 16 & 83.9 &\cmark\\
     {QKFormer~\cite{zhou2024qkformer}}
       & {6.74} & {4} & {96.18}
       & {6.74} & {4} & {81.15}
       & {1.50} & {16} & {98.6}
       & {1.50} & {16} & {84.0} &\xmark\\
     {{MS-QKFormer*}}
       & {6.74} & {4} & {96.84}
       & {6.74} & {4} & {81.57}
       & {1.50} & {16} & {98.6}
       & {1.50} & {16} & {82.3} &\cmark\\

     \midrule
     \textbf{Max-Former*}
       & \textbf{6.57} & \textbf{4} & \textbf{97.04}
       & \textbf{6.60} & \textbf{4} & \textbf{82.65}
       & \textbf{1.45} & \textbf{16} & \textbf{98.6}
       & \textbf{1.45} & \textbf{16} & \textbf{84.2} &\cmark\\
     \bottomrule
   \end{tabular}
   }
  }
\vspace{-0.3cm}
\end{table}
\begin{table}[b!]
\vspace{-0.1cm}
  \centering
  \caption{ImageNet performance comparison. Notation: A2S: ANN-to-SNN conversion; Model-L-D: models with L blocks and D channels. Input resolution: 224×224. \textsuperscript{*}Identical training configurations.}
  \label{tab:big_comparison}

  {\scriptsize
   \setlength{\tabcolsep}{4pt}      
   \renewcommand{\arraystretch}{0.8} 

   \resizebox{\textwidth}{!}{%
     \begin{tabular}{l c l c c c c c}
       \toprule
       \multirow{2}{*}{Methods}
         & \multirow{2}{*}{Type}
         & \multirow{2}{*}{Architecture}
         & \multirow{2}{*}{\makecell{Param\\(M)}}
         & \multirow{2}{*}{\makecell{Power\\(mJ)}}
         & \multirow{2}{*}{\makecell{Time\\Step}}
         & \multirow{2}{*}{\makecell{Top-1\\Acc (\%)}}
         & \multirow{2}{*}{\makecell{Membrane\\Shortcut}} \\
       & & & & & & & \\  
       \midrule
       ViT~\cite{vaswani2017attention}
         & ANN & ViT-L/16               &  304.3 & 80.96 & 1    & 79.70 & \textemdash \\
       \midrule
       DeiT~\cite{touvron2021training}
         & ANN & DeiT-B               & 86.6 & 80.50 & 1    & 81.80 & \textemdash \\
       \midrule

       PVT~\cite{wang2021pyramid}
         & ANN & PVT-Large               & 61.4 & 45.08 & 1    & 81.70 & \textemdash \\
       \midrule
       MST~\cite{wang2023masked}
         & A2S & Swin Transformer-T   & 28.50 & ---   & 512  & 78.51 & \xmark \\
       \midrule
       \multirow{2}{*}{Spikformer~\cite{zhou2022spikformer}}
         & \multirow{2}{*}{SNN}
         & Spikformer-8-384     & 16.81 & 7.73  & 4    & 70.24 & \xmark \\
         &     & Spikformer-8-768     & 66.34 & 21.48 & 4    & 74.81 & \xmark \\
       \midrule
       \multirow{2}{*}{S-Transformer~\cite{yao2023spike}}
         & \multirow{2}{*}{SNN}
         & S-Transformer-8-384   & 16.81 & 3.90  & 4    & 72.28 & \cmark \\
         &     & S-Transformer-8-768   & 66.34 & 6.10  & 4    &  76.30 & \cmark \\
       \midrule
       \multirow{2}{*}{Meta-Spikformer~\cite{yaospike}}
         & \multirow{2}{*}{SNN}
         & \textemdash   & 31.3 & 7.80  & 1    & 75.4 & \cmark \\
         & & \textemdash   & 31.3 & 32.80  & 4    & 77.2 & \cmark \\
       \midrule
       
       \multirow{1}{*}{SWformer~\cite{fang2024spiking}}
         & \multirow{1}{*}{SNN}
         & SWformer-8-512        & 27.6 & 5.08  & 4    & 75.43 & \cmark \\
       \midrule
       \multirow{2}{*}{{QKFormer~\cite{zhou2024qkformer}}}
         & \multirow{2}{*}{SNN}
         & HST-10-384            & 16.47 & 15.13 & 4    & 78.80 & \xmark \\
         &     & HST-10-768            & 64.96 & 8.52  & 1    & 81.69 & \xmark \\
       \midrule
       \multirow{2}{*}{{MS-QKFormer*}}
         & \multirow{2}{*}{SNN}
         & HST-10-384            & 16.47 & {5.52} & 4    & {76.48} & \cmark \\
         &     & HST-10-768            & 64.96 & {6.79}  & 1    & {77.78} & \cmark \\
       \midrule

       \multirow{5}{*}{\textbf{Max-Former*}}
         & \multirow{5}{*}{SNN}
         & Max-10-384            & 16.23 & \textbf{4.89} & 4    & \textbf{77.82} & \cmark \\
         &     & Max-10-512            & 28.65 & \textbf{2.50}  & 1    & \textbf{75.47} & \cmark \\
         &     & Max-10-512            & 28.65 & \textbf{7.49}  & 4    & \textbf{79.86} & \cmark \\
         &    & Max-10-768            & 63.99 & \textbf{5.27} & 1    & \textbf{78.60} & \cmark \\
         &     & Max-10-768            & 63.99 & \textbf{14.87}  & 4    & \textbf{82.39} & \cmark \\
         
       \bottomrule
     \end{tabular}%
   } 
  }
  \vspace{-0.1cm}
\end{table}

As shown in Table~\ref{tab:comparison}, Max-Former delivers performance improvements across both static datasets (CIFAR10/CIFAR100) and neuromorphic datasets (DVS128/CIFAR10-DVS). Notably, for CIFAR10/100 classification, its first stage only uses identity mapping for token mixing (Table~\ref{tab:configs}, yet still attains attractive results. Max-Former achieves 97.04\% accuracy on CIFAR10 with only 6.57M parameters at T=4, surpassing Spikformer (95.51\%, 9.32M), S-Transformer (95.60\%, 10.28M), and QKFormer (96.18\%, 6.74M). Similarly, on CIFAR100, Max-Former attains 82.65\% accuracy, significantly outperforming Spikformer (78.21\%), S-Transformer (78.40\%), and QKFormer (81.57\%).\vspace{5.5pt}

Max-Former and QKFormer share a similar hierarchical architecture, though QKFormer originally employs pre-spike shortcuts~\cite{zhou2024qkformer}. For a fair comparison, we additionally implemented QKFormer with the Membrane Shortcut (denoted as MS-QKFormer in the table) using identical training configurations. Max-Former still outperforms MS-QKFormer by 0.2\% on CIFAR10 (97.04\% vs. 96.84\%) and by 1.08\% on CIFAR100 (82.65\% vs. 81.57\%), while requiring slightly fewer parameters (6.57M vs. 6.74M). For neuromorphic datasets, Max-Former maintains this performance advantage. On DVS128, it achieves 98.6\% accuracy, matching MS-QKFormer with the membrane shortcut. On CIFAR10-DVS, Max-Former reaches 84.2\% accuracy, exceeding MS-QKFormer (82.3\%) by 1.9\% and surpassing other spike-driven models like S-Transformer (80.0\%) and SWformer (83.9\%).

       
\subsection{Results on ImageNet Classification}

Table~\ref{tab:big_comparison} shows Max-Former's performance on ImageNet classification, demonstrating its effectiveness for complex visual tasks.  Max-Former-10-768 (T=4) achieves 82.39\% accuracy (+7.58\% over Spikformer) with 30\% lower energy (14.87mJ vs 21.48mJ), despite using only lightweight for early-stage token mixing. It also outperforms the ANN-to-SNN MST model (78.51\%) that requires 512 timesteps. Training/inference speed and memory usage are analyzed in \textbf{Appendix~\ref{sec:a3}}.\vspace{5.5pt}

Our analysis focuses on models using the Membrane Shortcut, which eliminates the energy-inefficient ternary spike transmissions ($\{0,1,2\}$ in Pre-Spike Shortcut) while maintaining full compatibility with standard neuromorphic hardware without additional hardware overhead (see \textbf{Appendix~\ref{sec:a4}}).  For fair comparison, we implemented a membrane potential variant of QKFormer (denoted MS-QKFormer), where MS-QKFormer shows 64\% lower energy (5.52mJ vs 15.13mJ) in HST-10-384. \vspace{5.5pt}

Max-Former-10-384 (16.23M, T=4) achieves 77.82\% accuracy, outperforming MS-QKFormer-10-384 (16.47M, 76.48\%), S-Transformer-8-768 (66.34M, 76.3\%), and Meta-Spikformer (31.3M, 77.2\%). For energy efficiency under identical settings, Max-Former-10-384 consumes 4.89mJ, significantly lower than MS-QKFormer-10-384(5.52mJ), S-Transformer(6.10mJ), and  Meta-Spikformer (32.8mJ). Compared to conventional ANN models, Max-Former demonstrates concrete advantages in energy efficiency while maintaining competitive accuracy. Specifically, when compared to PVT-Large (a representative hierarchical ANN), Max-Former-10-768 (T=4) achieves comparable accuracy (82.34\% vs. 81.70\%) with 67\% lower energy consumption (14.87mJ vs. 45.08mJ). These results confirm the importance of high-frequency information in Spiking Transformers: replacing energy-intensive self-attention with lightweight DWC in early stages actually produces better performance.

\vspace{-0.4cm}
\begin{table}[ht]
  \centering
  \caption{Ablation of Patch-Embedding/ Token-Mixing Strategies on CIFAR100 and CIFAR10‐DVS.}
  \label{tab:ablation}

  \scriptsize
  \setlength{\tabcolsep}{6pt}
  \renewcommand{\arraystretch}{0.8}

  \resizebox{\textwidth}{!}{%
    \begin{tabular}{llllll}
      \toprule
      \multicolumn{3}{c}{\textbf{CIFAR100}} 
        & \multicolumn{3}{c}{\textbf{CIFAR10‐DVS}} \\
      \cmidrule(lr){1-3}\cmidrule(lr){4-6}
      Patch Embed                & Token Mix                         & Acc (\%) 
        & Patch Embed               & Token Mix              & Acc (\%) \\
      \midrule
      \textbf{Orig}/\textbf{Max}/\textbf{Max}
        & \textbf{Identity}/\textbf{DWC-3}/\underline{\textbf{SSA}}
        & \textbf{82.65}
        & \textbf{Max+}/\textbf{Max}
        & \textbf{DWC-3}/\textbf{SSA}
        & \textbf{84.2} \\
        \midrule
      Orig/\underline{\textbf{Orig}}/\underline{\textbf{Orig}}
        & Identity/DWC-3/SSA
        & 81.63
        & \underline{\textbf{Orig}}/\underline{\textbf{Orig}}
        & DWC-3/SSA
        & 79.2 \\
      Orig/Max/\underline{\textbf{Orig}}
        & Identity/DWC-3/SSA
        & 81.88
        & \underline{\textbf{Orig}}/Max
        & DWC-3/SSA
        & 81.5 \\
      Orig/Max/Max
        & Identity/\underline{\textbf{Identity}}/SSA
        & 81.28
        & Max+/Max
        & \underline{\textbf{DWC-1}}/SSA
        & 81.2 \\
      Orig/Max/Max
        & Identity/\underline{\textbf{DWC-5}}/SSA
        & 82.02
        & Max+/Max
        & \underline{\textbf{DWC-5}}/SSA
        & 82.7 \\
      Orig/Max/Max
        & \underline{\textbf{DWC-7}}/\underline{\textbf{DWC-5}}/SSA
        & 82.42
        & Max+/Max
        & \underline{\textbf{DWC-7}}/SSA
        & 82.1 \\
      \midrule
      Orig/Max/Max
        & \underline{\textbf{SSA}}/\underline{\textbf{SSA}}/SSA
        & 82.23
        & Max+/Max
        & \underline{\textbf{SSA}}/SSA
        & 83.9 \\
      Orig/Orig/Orig
        & \underline{\textbf{SSA}}/\underline{\textbf{SSA}}/SSA
        & 81.43
        & Orig/Orig
        & \underline{\textbf{SSA}}/SSA
        & 79.8 \\
      \bottomrule
    \end{tabular}%
  }
\vspace{-0.4cm}
\end{table}

\subsection{Ablation Study}

We provide direct evidence for the critical role of high-frequency information in Spiking Transformers through in-depth ablation studies.\vspace{5.5pt}

(1) \textbf{Ablation on Patch Embedding Strategies}: Proper patch embedding strategies help to unlock the performance limit of Spiking Transformers. On CIFAR100, changing Max-Former's patch embedding strategy from the proposed Embed-Orig/Embed-Max/Embed-Max to the default Embed-Orig/Embed-Orig/Embed-Orig configuration reduces performance from 82.65\% to 81.63\%. Neuromorphic datasets exhibit stronger dependence on high-frequency components. On CIFAR10-DVS, replacing the first stage's Embed-Max+ with Embed-Orig causes a significant performance drop from 84.2\% to 81.5\%.
Incorporating high-frequency information through patch embedding proves effective. When purely using SSA token mixing, optimizing the patch embedding strategy can improve accuracy by +0.8\% on Cifar 100 and +4.1\% on CIFAR10-DVS, highlighting its critical role in spiking architectures.\vspace{5.5pt}

(2) \textbf{Ablation on Token Mixing Strategies}: Max-Former further restores high-frequency information through the token mixing strategy. Despite SSA taking higher energy/parameter costs, Max-Former achieves better performance by simply replacing early-stage self-attention with lightweight DWC. On Cifar100, this substitution leads to +0.42\% performance gain (All SSA: 82.23\% vs. Max-Former: 82.65\%). On CIFAR10-DVS, Max-Former with hybrid token mixing (DWC-3 + SSA) achieves 84.2\% accuracy, outperforming the full-SSA variant (83.9\%) by 0.3\%. The kernel size selection for high-frequency preservation is also of great significance. Larger kernels (DWC-5/7) degrade performance (-0.63\% CIFAR100, -2.1\% CIFAR10-DVS) due to excessive feature smoothing. Insufficient filtering with DWC-1 also underperforms (81.2\% vs 84.2\% on CIFAR10-DVS). This highlights the necessity of balancing high-frequency and low-frequency components in Spiking Transformers.
\vspace{5.5pt}

Overall, Max-Former's performance gains stem from effective high-frequency information propagation, independent of parameter count, which is evidenced by:  (1) Max-pooling patch embedding (Embed-Max/Embed-Max+) consistently outperforms original versions despite similar parameter budgets. (2) Larger DWC kernels (DWC-5/DWC-7) increasing parameters but degrading accuracy (-0.63\% on CIFAR100, -2.1\% on CIFAR10-DVS vs Max-Former with DWC-3). See \textbf{Appendix~\ref{sec:a4}} for visualizations and\textbf{~\ref{sec:a5}} for limitations discussion.


\subsection{Generality across Convolutional Architectures}

We further extend the effectiveness of Max-Former to convolutional architectures by proposing Max-ResNet. The key modification in our Max-ResNet lies in the inclusion of only two additional max-pooling operations compared to MS-ResNet~\cite{hu2021advancing}. The detailed implementation of Max-ResNet is provided in Appendix~\ref{sec:max-resnet}. Training settings are listed in Appendix~\ref{sec:a1}. \vspace{0.2cm}

High-frequency information is essential for SNNs. As shown in Table~\ref{Tab:resnet}, Max-ResNet achieves a remarkable performance improvement over the baseline MS-ResNet, despite having identical model sizes.  Specifically, with the block configuration [2, 2, 2, 2], Max-ResNet improves the CIFAR-10 accuracy by +2.41\% (from 94.4\% to 96.81\%) and the CIFAR-100 accuracy by +6.48\%. Similarly, under the [3, 3, 2] configuration, the accuracy increases by +2.25\% and +6.65\% on CIFAR-10 and CIFAR-100, respectively. \vspace{0.2cm}

In short, Max‑ResNet‑18 sets state‑of‑the‑art benchmarks across convolutional baselines, with only a moderate model size and an extremely straightforward high-frequency restoration strategy.
Therefore, preserving high‑frequency information is fundamental to effective feature representation in SNNs, regardless of architecture.

\begin{table}[htbp]
\centering
\vspace{-0.4cm}
\caption{Comparison of different ResNet architectures on CIFAR-10 and CIFAR-100.}
\label{Tab:resnet}
\resizebox{\textwidth}{!}{
\begin{tabular}{l c c c c c c}
\toprule
Architecture & \makecell{Training\\Method} & \makecell{Block\\Config.} & \makecell{Params\\(M)} & \makecell{Time\\Step} & \makecell{CIFAR-10\\Acc. (\%)} & \makecell{CIFAR-100\\Acc. (\%)} \\
\midrule
KDSNN-ResNet-18~\cite{xu2023constructing} 
    & \makecell{Knowledge\\Distillation} & [2, 2, 2, 2] & 11.22 & 4 & 95.72 & 78.46 \\
\midrule
\multirow{2}{*}{MS-ResNet-18~\cite{hu2021advancing}} 
    & \multirow{3}{*}{\makecell{Direct\\Training}} & [2, 2, 2, 2] & 11.22 & 4 & 94.40 & 75.06 \\
    &  & [3, 3, 2] & 12.50 & 4 & 94.92 & 76.41 \\ \cmidrule(lr){1-1} \cmidrule(lr){3-7}
\multirow{1}{*}{MS-ResNet-34~\cite{hu2021advancing}} 
    &  & [2, 2, 2, 2] & 21.33 & 4 & 94.69 & 75.34 \\
\midrule
\multirow{2}{*}{Max-ResNet-18} 
    & \multirow{2}{*}{\makecell{Direct\\Training}} & [2, 2, 2, 2] & 11.22 & 4 & 96.81 (+2.41) & 81.54 (+6.48) \\
    &  & [3, 3, 2] & 12.50 & 4 & 97.17 (+2.25) & 83.06 (+6.65) \\
\bottomrule
\end{tabular}
}
\vspace{-0.6cm}
\end{table}

\section{Conclusion}
This work challenges the prevailing assumption that binary activation constraints are the primary cause of SNNs' performance gap. Through theoretical analysis and empirical validation, we demonstrate for the first time that spiking neurons inherently function as low-pass filters at the network level, resulting in the rapid attenuation of high-frequency components that critically degrade feature representation. \textbf{We demonstrate that high‑frequency information is crucial for effective spiking computation}: Max‑Former (63.99M parameters) achieves 82.39\% top‑1 accuracy on ImageNet, outperforming Spikformer (74.81\%, 66.34M parameters) by +7.58\%, while reducing energy consumption by 30\% at a comparable model size; Max‑ResNet‑18 further achieves state‑of‑the‑art performance among convolutional baselines -- 97.17\% on CIFAR‑10 and 83.06\% on CIFAR‑100. \textbf{Notably, all these improvements are achieved through extremely simple modifications, even slightly reducing the overall model size.}  We believe this simple yet effective solution will motivate future research to explore the unique properties of SNNs, beyond the established practices in ANN studies.


\section{Acknowledgements}
This work is supported by the Major Science and Technology Innovation 2030 “Brain Science and Brain-like Research” key project(No.2021ZD0201405), the Guangzhou-HKUST(GZ) Joint Funding Program (Grant No. 2023A03J0682), the National Natural Science Foundation of China (Grant No. 62405255), GuangDong Basic and Applied Basic Research Foundation (No. 2023A1515110679), and partially supported by the collaborative project with Brain Mind Innovation, Inc

\bibliographystyle{unsrt}
\bibliography{ref}
\medskip


\newpage
\appendix

\section{Analysis of High-Frequency Information in Spiking Transformers}
\label{sec:a1}

To validate our approach, we conduct experiments across three static and two neuromorphic datasets. In this section, we first provide the detailed experimental setup used to obtain the results presented in our paper. We then perform additional analysis on the importance of high-frequency information in Spiking Transformers. For complete parameter configurations, please refer to our public code repository at \href{https://github.com/bic-L/MaxFormer}{https://github.com/bic-L/MaxFormer}.

\begin{table}[ht]
\centering
\vspace{-0.2cm}
\caption{Hyperparameters for image classification on different datasets. }
\vspace{0.2cm}
\label{tab:hyperparameters}
\begin{tabular}{@{} l c c c c @{}}
\toprule
\textbf{Hyper parameters}    & \textbf{ImageNet}                                 & \textbf{CIFAR-10}       & \textbf{CIFAR-100}      & \textbf{Neuromorphic} \\ 
\midrule
Model Size         & 10--384 / 10--512 / 10--768              & 4--384         & 4--384         & 2--256       \\ \midrule
Epochs             & 200                                      & 400            & 400            & 106          \\ \midrule
Resolution         & $224\times224$                           & $32\times32$   & $32\times32$   & $128\times128$ \\ \midrule
Batch Size         & 512 (8 $\times$ 64)                                      & 128            & 64             & 16           \\ \midrule
Optimizer          & AdamW                                    & AdamW          & AdamW          & AdamW        \\
\midrule
Learning rate      & \makecell[c]{$1.2\times10^{-3}\,(T=1)$ \\ $1.35\times10^{-3}\,(T=4)$}
                    & \makecell[c]{$1.50\times10^{-3}$}
                    & \makecell[c]{$1.50\times10^{-3}$}
                    & \makecell[c]{$6.00\times10^{-3}$} \\
\midrule
Learning rate decay& Cosine                                   & Cosine         & Cosine         & Cosine       \\ \midrule
Warmup epochs      & 5                                        & 20             & 20             & 10           \\ \midrule
Weight decay       & 0.05                                     & 0.06           & 0.06           & 0.06         \\ \midrule
Rand Augment       & 9 / 0.5                                  & 9-n1 / 0.4        & 9-n1 / 0.4    & ---          \\ \midrule
Mixup              & 0.25~/~0.4/~0.8              & 0.5            & 0.75           & 0.5          \\ \midrule
CutMix             & 1                                        & 0.5            & 0.5            & ---          \\ \midrule
Mixup prob         &  0.5                        & 1              & 1              & 0.5          \\ \midrule
Erasing prob       & 0.0                                        & 0.25           & 0.25           & ---          \\ \midrule
Label smoothing    & 0.1                                      & 0.1            & 0.1            & 0.1          \\
\bottomrule
\end{tabular}
\end{table}
\vspace{-0.2cm}

\subsection{Experimental Details}
\vspace{5.5pt}

\textbf{Datasets}: We evaluate Max-Former through comprehensive experiments on static datasets (CIFAR-10~\cite{lecunGradientbasedLearningApplied1998a}, CIFAR-100~\cite{krizhevskyLearningMultipleLayers2009a} and ImageNet~\cite{dengImagenetLargescaleHierarchical2009}) and neuromorphic datasets (CIFAR10-DVS~\cite{liCIFAR10DVSEventStreamDataset2017}, DVS128 Gesture~\cite{amir2017low}). The training and inference pipeline are implemented in SpikingJelly~\cite{SpikingJelly}.
\vspace{5.5pt}

Static Datasets: For static image classification, we evaluate on three standard benchmarks. {ImageNet-1k}~\cite{dengImagenetLargescaleHierarchical2009} is one of the most widely used datasets in computer vision. It contains 1.28 million training, 50,000 validation, and 100,000 test images covering the common 1K classes. Both CIFAR-10~\cite{lecunGradientbasedLearningApplied1998a} and CIFAR-100~\cite{krizhevskyLearningMultipleLayers2009a} include 50,000 training images and 10,000 testing images with 32$\times$32 resolution. The main difference between them is that CIFAR-10 has 10 categories for classification, while CIFAR-100 has 100 categories.
\vspace{5.5pt}

Neuromorphic Datasets: For event-based vision tasks, we evaluate on two standard benchmarks. CIFAR10-DVS~\cite{liCIFAR10DVSEventStreamDataset2017} is an event-based version of the CIFAR-10 dataset, created by capturing moving image samples using the Dynamic Vision Sensor (DVS). It includes 10,000 event-based images (128$\times$128 pixels) spread across 10 classes, with 9,000 samples for training and 1,000 for testing. The DVS128 Gesture dataset~\cite{amir2017low} contains 1,342 event-based recordings of 11 different hand gesture types performed by 29 people under 3 different lighting conditions. Each gesture recording lasts about 6 seconds on average.
\vspace{5.5pt}

\textbf{Hyper Parameters: } Our training scheme mainly follows~\cite{zhou2024qkformer} and ~\cite{yao2023spike}. Specifically, MixUp~\cite{zhang2018mixup}, CutMix~\cite{yun2019cutmix} and  RandAugment~\cite{cubuk2020randaugment} are used for data augmentation. The models are trained using AdamW optimizer~\cite{loshchilov2017decoupled} with the weight decay of 0.05 for ImageNet-1K classification tasks and the weight decay of 0.06 for all other datasets. Label Smoothing~\cite{szegedy2016rethinking} is set as 0.1. Detailed training hyperparameters are shown in Table~\ref{tab:hyperparameters}. For our ImageNet experiments, we used 8 NVIDIA A30 GPUs to train most models. However, for the MaxFormer-10-512 (T=4) and MaxFormer-10-768 (T=4) models, we used 8 NVIDIA H20 GPUs instead. For the smaller datasets (CIFAR10, CIFAR100, DVS128 Gesture, and CIFAR10-DVS), we used a single A30 GPU for training.
 \vspace{5.5pt}

\subsection{Impact of High-Frequency Information on Model Performance}

\begin{table}[ht]
\vspace{-0.5cm}
  \centering
  \caption{Patch embedding and token mixing schemes on CIFAR-100. Acc.:Top-1 Accuracy (\%). DWC-$N$: spiking depth-wise convolution with kernel size $N \times N$. SSA: Spiking Self-Attention.}
  \vspace{0.2cm}
  \label{tab:patch-token-cifar100}
  \small
  \begin{tabular}{@{} c l l c @{}} 
    \toprule
          & \textbf{Patch Embed} & \textbf{Token Mix}                  & \textbf{Acc. (\%)} \\ 
    \midrule
  \multirow{2}{*}{(1)} 
          & \multirow{2}{*}{Embed-Orig/Embed-Orig/Embed-Orig}               
                                                & Avg-Pool/Avg-Pool/Avg-Pool         & 76.73 \\
          &                                                & Max-Pool/Max-Pool/Max-Pool         & 79.12 \\
    \midrule

  \multirow{2}{*}{(2)} 
          & Embed-Orig/Embed-Orig/\textbf{Embed-Max}       
                                                & Identity/Identity/Identity         & 80.11 \\
          & Embed-Orig/\textbf{Embed-Max}/\textbf{Embed-Max} 
                                                & Identity/Identity/Identity         & 80.46 \\
    \midrule
  \multirow{4}{*}{(3)} 
          & \multirow{4}{*}{Embed-Orig/\textbf{Embed-Max}/\textbf{Embed-Max}} 
                                                & Avg-Pool/Avg-Pool/Avg-Pool         & 77.61 \\
          &                                                & Max-Pool/Max-Pool/Max-Pool         & 79.78 \\
           &                                     & Identity/Max-Pool/Identity         & 79.99 \\
          &                                                & Identity/Identity/Max-Pool         & 80.12 \\
    \midrule
  \multirow{7}{*}{(4)} 
          &  \multirow{7}{*}{Embed-Orig/\textbf{Embed-Max}/\textbf{Embed-Max}} 
                                                & SSA/SSA/SSA                        & 82.13 \\
          &                                                & DWC-3/DWC-5/SSA                    & 82.36 \\
          &                                                & DWC-5/DWC-3/SSA                    & 82.46 \\
          &                                                & DWC-5/DWC-7/SSA                    & 82.45 \\
          &                                                & DWC-7/DWC-5/SSA                    & 82.42 \\
          &                                                & DWC-3/DWC-3/SSA                    & 82.59 \\
          &                                                & Identity/DWC-3/SSA                 & 82.65 \\
    \midrule
  \multirow{3}{*}{(5)} 
          & Embed-Orig/\textbf{Embed-Max}/\textbf{Embed-Max} 
                                                & SSA+DWC-5/SSA+DWC-5/SSA+DWC-5         & 82.09 \\
          & Embed-Orig/Embed-Orig/Embed-Orig            
                                                & SSA+DWC-3/SSA+DWC-3/SSA+DWC-3         & 82.56 \\
          & Embed-Orig/Embed-Orig/Embed-Orig            
                                                & SSA+DWC-5/SSA+DWC-5/SSA+DWC-5         & 82.73 \\
                \midrule
  \multirow{3}{*}{(6)} 
          & Embed-Orig/Embed-Max/Embed-Max       & SSA/SSA/SSA         & 82.13 \\
          & Embed-1Max/Embed-Max/Embed-Max            
                                                & SSA/SSA/SSA         & 79.98 \\
          & Embed-Max/Embed-Max/Embed-Max            
                                                & SSA/SSA/SSA         & 78.78 \\
                \midrule
  \multirow{2}{*}{(7)} 
          & Embed-Orig/Embed-1Max/Embed-1Max 
                                                & SSA/SSA/SSA         & 82.02 \\
          & Embed-1Max/Embed-1Max/Embed-1Max	            
                                                & SSA/SSA/SSA         & 81.86 \\
                \midrule
  \multirow{2}{*}{(8)} 
          & Embed-Orig/Embed-Max/Embed-Max            
                                                & SDSA/SDSA/SDSA         & 81.77 \\
          & Embed-Orig/Embed-1Max/Embed-1Max            
                                                & SDSA/SDSA/SDSA         & 81.52 \\                                                
    \bottomrule
  \end{tabular}
\vspace{5.5pt}
\end{table}

In Table~\ref{tab:patch-token-cifar100}, we provide additional analysis on the impact of high-frequency information. We conducted these experiments on CIFAR-100~\cite{krizhevskyLearningMultipleLayers2009a}, using the experimental settings detailed in Table~\ref{tab:hyperparameters}. We discuss the ablation below according to the following aspects: \vspace{6.5pt}

\textbf{(a)~Extra Max-Pool in Patch Embedding/Token Mixing: }
\vspace{5.5pt}

\textbf{High-frequency information plays a critical role in the performance of Spiking Transformer due to the inherent low-pass filter characteristics of spiking neurons.} Experimental results shown in Table~\ref{tab:patch-token-cifar100}~(1) reveal that strategically preserving these frequencies through max-pooling operations significantly enhances model accuracy, with a 2.39\% improvement when replacing average pooling with max-pooling across all stages (76.73\% to 79.12\%). In Table~\ref{tab:patch-token-cifar100} (2), the performance of Spiking Transformer increases progressively when extending Embed-Max from the last patch embedding blocks (80.11\%) to include the middle block (80.46\%).
\vspace{5.5pt}

\textbf{However, excess high-frequency information will instead impair model performance.} For instance, in Table~\ref{tab:patch-token-cifar100}~(3), switching from using avg-pool for all token mixing to max-pool improves top-1 accuracy from 77.61\% to 79.78\%. Yet we found even better results with a more targeted setting: when employing max-pooling exclusively in the middle stage increases accuracy to 79.99\%, while restricting it to only the last stage further pushes performance to 80.12\%.
\vspace{5.5pt}

This happens because spiking neurons act like low-pass filters that naturally reduce high-frequency components as information moves deeper through the network. Therefore, strategically adding back high-frequency components at specific points in the network is crucial for pushing the performance limit of Spiking Transformers.
\vspace{5.5pt}

 In general, \textbf{for the hierarchical spiking transformer, Embed-Max can improve performance when used for patch embedding}, as shown in Table~\ref{tab:patch-token-cifar100}~(6-8). However, for non-hierarchical ones, although enhancing high-frequency components can in principle be beneficial, its actual impact is limited since patch embedding is only performed once throughout the entire process, causing the improvement to diminish in deeper layers. A more promising solution lies in optimizing token mixing, as discussed in~\cite{fang2024spiking}.\vspace{5.5pt}


\textbf{(b)~Spiking Transformers Benefit from High-Frequency Information: }
\vspace{5.5pt}

In biological vision, high-frequency details help early processing stages learn elementary features,, which are then gradually built from local to global representations. Similarly, in standard non-spiking Transformers, the lower layers typically need more high-frequency details, while higher layers work better with global information. Spiking Transformers follows the same design philosophy, but with an important difference: they need additional frequency-enhancing operations (e.g., max-pooling and depth-wise convolution) for restoring high-frequency information that would otherwise be lost.
\vspace{5.5pt}

In Table~\ref{tab:patch-token-cifar100}~(4), we show that, a proper token mixing strategy can effectively restore high-frequency information in Spiking Transformer, resulting in significant performance gains. By replacing Spiking Self-Attention (SSA) with depth-wise convolution (DWC) for token mixing, we improved performance from 82.13\% to 82.65\%.  Importantly, this improvement does not come from adding more parameters/ computational burden. For example, the Identity/DWC-3/SSA combination works 0.29\% better than DWC-3/DWC-5/SSA, even though the former one has a lower computational cost. Our further experiments in Table~\ref{tab:patch-token-cifar100}~(4-5) confirm that these findings hold true in the full-SSA network: restoring high-frequency components can significantly optimize the performance from 82.13\% to 82.73\% (+0.6\%) on Cifar100. The proper high-frequency enhancement strategy is essential to unlocking the full potential of Spiking Transformers.
\vspace{5.5pt}

\subsection{Max-ResNet Implementation}
\label{sec:max-resnet}

\begin{figure*}[ht]
\includegraphics[width=1.0\linewidth]
{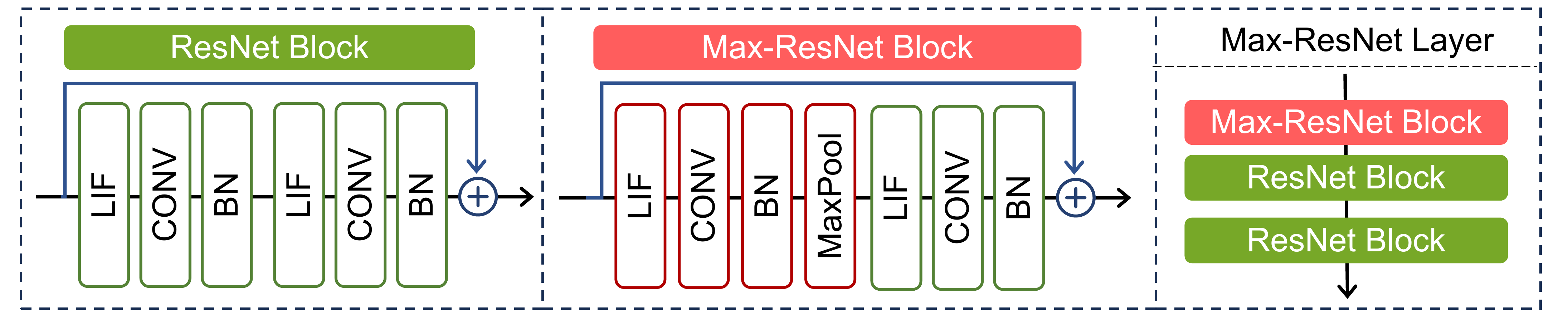}
\caption{Overview of Max-ResNet. A single Max‑Pool operation is added per block and layer.}
\label{fig:max-resmet}
\end{figure*}

As shown in Figure~\ref{fig:max-resmet}, Max‑ResNet introduces only a minor architectural change to MS‑ResNet~\cite{hu2021advancing}: all layers are replaced with Max‑ResNet layers, while the first layer remains unchanged. Code implementation is available at \href{https://github.com/bic-L/MaxFormer}{https://github.com/bic-L/MaxFormer}.

\section{Energy Analysis}
\label{sec:a2}

To estimate the theoretical energy consumption of Spiking Transformers, we follow the methodology used in previous studies~\cite{zhou2022spikformer, hu2021spiking, yao2023spike, fang2024spiking}. It is worth noting that the batch normalization (BN) layers and linear scaling transformations that follow convolution layers can be combined directly into the convolution layers themselves with added bias terms during deployment. Thus, in common practice~\cite{zhou2022spikformer, hu2021spiking, yao2023spike, fang2024spiking}, the energy consumption of BN is typically excluded when calculating theoretical energy usage. For fair comparison, our work adopts the same strategy.  In Spiking Transformers, energy consumption is directly proportional to synaptic operations (SOPs), which can be calculated as:
\vspace{2.5pt}
\begin{equation}
\text{SOPs}(l) = fr \times \text{T} \times \text{FLOPs}(l)
\end{equation}


\noindent where $l$ represents a specific block or layer in the Spiking Transformer architecture, $fr$ refers to the firing rate of the input spike train for that particular block or layer, and $\text{T}$ is the simulation time step of spiking neurons. Assuming the multiply-accumulate (MAC) and accumulate (AC) operations are implemented on the 45nm neuromorphic chip described in~\cite{davies2018loihi}, where each MAC operation uses $E_\text{MAC}$ = 4.6pJ of energy and each AC operation uses $E_\text{AC}$ = 0.9pJ, we can estimate the total energy consumption of a Spiking Transformer by adding up the energy used by all MAC and AC operations across all layers:
\vspace{2.5pt}
\begin{equation}
\begin{aligned}
    E_\text{SNN} &= E_\text{MAC} \times \text{FLOP}^{{1}}_\text{CONV}
    + E_\text{AC} \times (\sum_{n=2}^{N}\text{SOP}^{n}_\text{SNN Conv} + \sum_{j=1}^{M}\text{SOP}^{j}_\text{SNN FC})
\end{aligned}
\end{equation}

$\text{FLOP}^{{1}}_\text{CONV}$ represents the floating-point operations in the first layer, which converts non-spike inputs into spike form for static image classification tasks. Since this layer performs floating-point computations, we estimate its energy consumption using $E_\text{MAC}$. For all subsequent layers, which process spike data, we estimate energy consumption using $E_\text{AC}$. For mainstream non-spiking Transformers, the energy consumption is estimated through:

\begin{equation}
\begin{aligned}
    E_\text{ANN} &= E_\text{MAC} \times \text{FLOPs}
\end{aligned}
\end{equation}

\section{Comparison on Train/Inference Time and Memory Consumption}
\label{sec:a3}

\begin{table}[ht]
\vspace{-0.5cm}
  \centering
  \small
  \caption{Comparison of training/inference time and memory usage between QKFormer-10-768 and Max-Former-10-768 models. All measurements were conducted with the simulation timestep of 1 and the batch size of 32. MS-QKFormer indicates the QKFormer variant with the membrane shortcut.}
  \vspace{0.2cm}
  \label{tab:time-memory}
  \resizebox{\textwidth}{!}{%
    \begin{tabular}{@{} l c c c c @{}}
      \toprule
      & \makecell{Train \\ Time (s)} 
      & \makecell{Train \\ Memory (MB)} 
      & \makecell{Infer.\\ Time (s)} 
      & \makecell{Infer.\\ Memory (MB)} \\
      \midrule
      QKFormer (64.96M, T=1, B=32)    & 0.214 & 18227 & 0.053 & 5000 \\
      MS-QKFormer* (64.96M, T=1, B=32)   & 0.208 & 17496 & 0.048 & 4822 \\
      Max-Former* (63.99M, T=1, B=32)  & 0.179 & 15431 & 0.044 & 4354 \\
      \bottomrule
    \end{tabular}%
  }
  \vspace{0.2cm}
\end{table}

Max-Former delivers faster training and inference speeds while consuming less memory. We compared the performance metrics of QKFormer~\cite{zhou2024qkformer}, its membrane shortcut variant (MS-QKFormer), and Max-Former on ImageNet using 224 $\times$ 224 input resolution. All tests were conducted on a CentOS 7.9 server equipped with the Intel Xeon Gold 6348 CPU (2.60GHz) and the Nvidia A30 GPU. As Table~\ref{tab:time-memory} shows, when compared to MS-QKFormer with the same hierarchical architecture and shortcut configuration, Max-Former reduces training time by 14\% and both inference time and memory usage by 10\%. Additionally, our results indicate that the pre-spike shortcut strategy used in the original QKFormer increases both processing time and memory demands.

\section{Residual Connections in Spiking Neural Networks}
\label{sec:a4}

\textbf{(a) Performance and Energy Tradeoffs:}
\vspace{5.5pt}

The unique asynchronous nature of spike-based computation makes implementing residual connections challenging.  As a result, the research community of spiking neural networks (SNNs) has not yet reached a consensus on the standardized residual learning approach, either in terms of algorithm or hardware implementation. While the main focus of our work is not related to residual learning, we want to offer a detailed comparison between the pre-spike shortcut~\cite{fang2021deep} and the membrane shortcut~\cite{hu2021spiking}, the two most representative residual learning methods that emerged in recent years.\vspace{5.5pt}

As explained in Section \textbf{3.3}, the pre-spike shortcut~\cite{fang2021deep} implements residual connections between spiking outputs, while the membrane shortcut~\cite{hu2021spiking} connects membrane potentials directly.  In algorithm designs, membrane shortcuts have been reported to facilitate better performance in prior works~\cite{zhou2024qkformer, yao2023spike}, especially on small datasets. However, our findings indicate this advantage is not universal across all scenarios. 
As shown in Table~\ref{tab:energy}, the patch embedding stages of Spiking Transformers account for the majority of energy consumption. Consequently, the multiple patch embedding stages in hierarchical architectures like QKFormer ~\cite{zhou2024qkformer}, while enabling efficient feature learning with fewer parameters, also come at higher energy usage. This makes the choice of shortcut scheme particularly impactful on overall energy efficiency. When processing ImageNet images at 224 × 224 resolution, the pre-spike QKFormer consumes three times more energy than its membrane shortcut variant, reflecting the substantially higher SOPs required. Nevertheless, this increased computational overhead does translate to notable performance gains (+2.32\% accuracy when comparing QKFormer to MS-QKFormer).\vspace{5.5pt}

From the hardware perspective, implementing either shortcut type on neuromorphic chips is technically feasible but presents significant challenges~\cite{weidel2021wavesense, balafrej2025enhancing}. Implementing the pre-spike shortcut specifically requires the chips to support multi-spike operations, as the ternary spike transmissions (values of 0, 1, or 2) can occur. This results in either higher energy consumption or increased hardware complexity~\cite{gautam2023adaptive}.
Yao et al.~\cite{yao2023spike} proposed implementing membrane shortcuts through the addressing function that passes the membrane potential to corresponding neurons in subsequent layers for merging. While membrane shortcuts do strictly adhere to the spike-driven computing paradigm and could possibly be supported by standard neuromorphic hardware, transmitting membrane potentials can create substantial communication overhead, making practical implementation non-trivial. Many discussions on hardware deployment still advocate avoiding shortcuts as the current preferred approach~\cite{deng2024spiking, meyer2024diagonal}. However, given the critical role of residual learning in modern deep learning, avoiding shortcuts altogether is not a sustainable long-term strategy for advancing neuromorphic computing.\vspace{5.5pt}

In our work, we primarily compare with MS-QKFormer using the membrane shortcut to ensure fair comparisons.
We hope the SNN community can establish a consensus on standardized shortcut implementations in the near future, considering the significant impact of shortcut schemes on both energy efficiency and model performance.

\begin{table}[ht]
    \vspace{-0.2cm}
  \centering
  \small
  \setlength{\tabcolsep}{2pt}
  \caption{ Energy consumption comparison (in mJ) across processing stages for QKFormer, QKFormer with membrane shortcut (MS-QKFormer), and our proposed Max-Former model. * Identical training configurations.}
  \vspace{0.2cm}
  \label{tab:energy}
\resizebox{\textwidth}{!}{%
  \begin{tabular}{@{} l
                   ccc   
                   ccc   
                   ccc   
                   c     
                   c     
                @{}}
    \toprule
    \makecell{\multirow{4}{*}{Model}}
      & \multicolumn{3}{c}{Stage 1}
      & \multicolumn{3}{c}{Stage 2}
      & \multicolumn{3}{c}{Stage 3}
      & \multirow{4}{*}{\makecell{Classifier}}
      & \multirow{4}{*}{\makecell{Total\\Energy}} \\
    \cmidrule(lr){2-4} \cmidrule(lr){5-7} \cmidrule(lr){8-10}
    
      &  \makecell{Patch\\Embed} & \makecell{Token\\Mix} & MLP
      & \makecell{Patch\\Embed} & \makecell{Token\\Mix} & MLP
      & \makecell{Patch\\Embed} & \makecell{Token\\Mix} & MLP
      &                 
      &                 \\
    \midrule
    QKFormer (16.47M, T=4, 78.80\%)
      & 1.26  & 0.16  & 0.41  & 2.68  & 0.35  & 0.77  
      & 2.97  & 2.71  & 3.81  & 0.006 & 15.13 \\
    MS-QKFormer* (16.47M, T=4, 76.48\%)
      & 1.19  & 0.052 & 0.15  & 0.96  & 0.11  & 0.25  
      & 0.88  & 0.88  & 1.06  & 0.007 &  5.52 \\
    Max-Former* (16.23M, T=4, 77.82\%)
      & 0.41  & 0.02  & 0.17  & 0.89  & 0.01  & 0.36  
      & 0.91  & 0.96  & 1.16  & 0.001 &  4.89 \\
    \bottomrule
  \end{tabular}
  }
  \vspace{-0.5cm}
\end{table}

\vspace{5.5pt}
\section{Visualization}
\label{sec:a5}
\vspace{5.5pt}

We demonstrate the GradCAM visualizations~\cite{selvarajuGradcamVisualExplanations2017} of four Spiking Transformers with the membrane shortcut of similar size. Compared to Spike-Driven Transformer~\cite{yao2023spike} and SWFormer~\cite{fang2024spiking}, the hierarchical architecture used in both QKFormer~\cite{zhou2024qkformer} and our Max-Former allows for more precise focusing on target objects. Compared to MS-QKFormer, Max-Former shows more concentrated activation patterns. For instance, in the polar bear image, Max-Former completely skips the background and precisely focuses on the bear's key features (the head, rather than the outline or fur).

\begin{figure*}[ht]
\includegraphics[width=1.0\linewidth]
{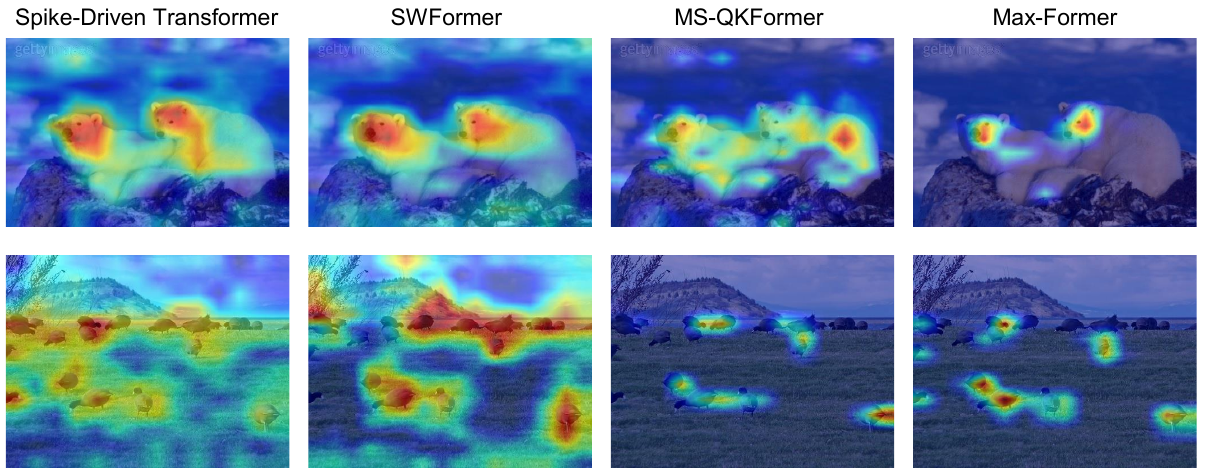}
\caption{GradCAM visualizations~\cite{selvarajuGradcamVisualExplanations2017} comparing four Spiking Transformers of similar size: Spike-Driven Transformer-8-512~\cite{yao2023spike} (29.68M), SWFormer-8-512~\cite{fang2024spiking} (27.6M), QKFormer-10-512 with membrane shortcut (MS-QKFormer) (29.08M), and our Max-Former-10-512 (28.65M).}
\label{fig:cam}
\end{figure*}

\section{Impact and Limitation}
\label{sec:a6}

Our work serves as the theoretical foundation for many existing architectural choices in Spiking Transformers. Specifically, in ~\cite{yao2023spike}, the authors found that directly applying known practices of MetaFormer does not achieve good results in Spiking Transformers: employing average pooling operators to replace SDSA as the token mixer surprisingly results in substantial performance degradation from 61.0\% to 41.2\%. Similar phenomena have been discussed in earlier works. For instance, Spikformer v2~\cite{zhou2024spikformer} discovered that removing the max-pooling operator in the original Spikformer~\cite{zhou2022spikformer} leads to a substantial performance drop, while adding convolution layers (which act as high-pass filters) in the patch embedding stage significantly improves the performance. Our work reveals the underlying principles of these architectural designs: spiking transformers need to enhance high-frequency components to alleviate the feature degradation caused by their inherent low-pass activation. \vspace{0.05cm}

We are well aware that there is still much space to be explored, and we hope that our Max-Former can serve as a good starting point for future research. For instance, similar to~\cite{si2022inception}, Max-Former requires manually balancing frequency components, which demands considerable expertise when adapting to different tasks. Incorporating direct frequency learning approaches like Fourier-based~\cite{li2020fourier} or Wavelet-based\cite{fang2024spiking} methods would offer more straightforward solutions. The main challenge, however, lies in developing efficient spike-based frequency representations. Overall, we believe our work will inspire more future research to advance neuromorphic computing through exploring the unique properties of spiking neurons, rather than expending excessive effort to adapt established practices from standard non-spiking neural networks.



\newpage
\section*{NeurIPS Paper Checklist}


\begin{enumerate}

\item {\bf Claims}
    \item[] Question: Do the main claims made in the abstract and introduction accurately reflect the paper's contributions and scope?
    \item[] Answer: \answerYes{} 
    \item[] Justification: The abstract and introduction reflect the paper's contributions and scope clearly.

\item {\bf Limitations}
    \item[] Question: Does the paper discuss the limitations of the work performed by the authors?
    \item[] Answer: \answerYes{} 
    \item[] Justification: We discuss the limitations in \textbf{Appendix~\ref{sec:a5}}.

\item {\bf Theory assumptions and proofs}
    \item[] Question: For each theoretical result, does the paper provide the full set of assumptions and a complete (and correct) proof?
    \item[] Answer: \answerYes{} 
    \item[] Justification: All the theorems, formulas, and proofs are clearly stated in the main paper.

    \item {\bf Experimental result reproducibility}
    \item[] Question: Does the paper fully disclose all the information needed to reproduce the main experimental results of the paper to the extent that it affects the main claims and/or conclusions of the paper (regardless of whether the code and data are provided or not)?
    \item[] Answer:  \answerYes{}
    \item[] Justification: We provide experiment details in the \textbf{Appendix~\ref{sec:a1},~\ref{sec:a2}}.

\item {\bf Open access to data and code}
    \item[] Question: Does the paper provide open access to the data and code, with sufficient instructions to faithfully reproduce the main experimental results, as described in supplemental material?
    \item[] Answer: \answerYes{} 
    \item[] Justification: \href{https://github.com/bic-L/MaxFormer}{Code is available: https://github.com/bic-L/MaxFormer}.

\item {\bf Experimental setting/details}
    \item[] Question: Does the paper specify all the training and test details (e.g., data splits, hyperparameters, how they were chosen, type of optimizer, etc.) necessary to understand the results?
    \item[] Answer:  \answerYes{} 
    \item[] Justification: All training and test details have been revealed in the \textbf{Appendix~\ref{sec:a1}}.

\item {\bf Experiment statistical significance}
    \item[] Question: Does the paper report error bars suitably and correctly defined or other appropriate information about the statistical significance of the experiments?
    \item[] Answer: \answerNo{}
    \item[] Justification: Error bars are not reported because it would be too computationally expensive.

\item {\bf Experiments compute resources}
    \item[] Question: For each experiment, does the paper provide sufficient information on the computer resources (type of compute workers, memory, time of execution) needed to reproduce the experiments?
    \item[] Answer: \answerYes{} 
    \item[] Justification: See \textbf{Appendix~\ref{sec:a1}}.

\item {\bf Code of ethics}
    \item[] Question: Does the research conducted in the paper conform, in every respect, with the NeurIPS Code of Ethics \url{https://neurips.cc/public/EthicsGuidelines}?
    \item[] Answer: \answerYes{}
    \item[] Justification:  This research is in every respect with the NeurIPS Code of Ethics.

\item {\bf Broader impacts}
    \item[] Question: Does the paper discuss both potential positive societal impacts and negative societal impacts of the work performed?
    \item[] Answer: \answerNA{} 
    \item[] Justification: This research belongs to foundational research and is not tied to particular applications.
    
\item {\bf Safeguards}
    \item[] Question: Does the paper describe safeguards that have been put in place for responsible release of data or models that have a high risk for misuse (e.g., pretrained language models, image generators, or scraped datasets)?
    \item[] Answer: \answerNA{} 
    \item[] Justification: The paper poses no such risks.

\item {\bf Licenses for existing assets}
    \item[] Question: Are the creators or original owners of assets (e.g., code, data, models), used in the paper, properly credited and are the license and terms of use explicitly mentioned and properly respected?
    \item[] Answer: \answerYes{} 
    \item[] Justification: The paper properly credits creators and mentions the license and terms of use for existing assets.

\item {\bf New assets}
    \item[] Question: Are new assets introduced in the paper well documented and is the documentation provided alongside the assets?
    \item[] Answer: \answerNA{} 
    \item[] Justification:  The paper does not release new assets.

\item {\bf Crowdsourcing and research with human subjects}
    \item[] Question: For crowdsourcing experiments and research with human subjects, does the paper include the full text of instructions given to participants and screenshots, if applicable, as well as details about compensation (if any)? 
    \item[] Answer: \answerNA{} 
    \item[] Justification: The paper does not involve crowdsourcing or research with human subjects.

\item {\bf Institutional review board (IRB) approvals or equivalent for research with human subjects}
    \item[] Question: Does the paper describe potential risks incurred by study participants, whether such risks were disclosed to the subjects, and whether Institutional Review Board (IRB) approvals (or an equivalent approval/review based on the requirements of your country or institution) were obtained?
    \item[] Answer: \answerNA{} 
    \item[] Justification: The paper does not involve crowdsourcing or research with human subjects.

\item {\bf Declaration of LLM usage}
    \item[] Question: Does the paper describe the usage of LLMs if it is an important, original, or non-standard component of the core methods in this research? Note that if the LLM is used only for writing, editing, or formatting purposes and does not impact the core methodology, scientific rigorousness, or originality of the research, declaration is not required.
    \item[] Answer: \answerNA{} 
    \item[] Justification: The paper does not involve LLMs as any important, original, or non-standard components.

\end{enumerate}

\end{document}